\documentclass[a4paper,onesided,12pt]{report}
\pdfoutput=1
\usepackage{styles/fbe_tez}
\usepackage[utf8x]{inputenc} 

\usepackage{amsmath, amsthm, amssymb}
\usepackage[bottom]{footmisc}
\usepackage{cite}
\usepackage{graphicx}
\usepackage{graphicx}
\usepackage{longtable}
\usepackage{lscape}
\usepackage{floatpag}
\usepackage{calrsfs}
\usepackage{amsmath}
\usepackage{rotating}
\usepackage{todonotes}
\usepackage[all]{nowidow}
\graphicspath{{images/}} 
\usepackage{xspace}
\newcommand{\bftab}{\fontseries{b}\selectfont}

\usepackage{multirow}
\usepackage{subfigure}
\usepackage{algorithm}
\usepackage{algorithmic}
\usepackage{chngcntr}
\counterwithin{table}{chapter}
\DeclareMathAlphabet{\pazocal}{OMS}{zplm}{m}{n}

\newcommand{\CommaPunct}{\mathpunct{\raisebox{0.5ex}{,}}}
\newcommand{\latinphrase}[1]{\textit{#1}} 
\newcommand{\etal}{\latinphrase{et~al.}\xspace}

\title{NEURAL SIGN LANGUAGE TRANSLATION BY LEARNING TOKENIZATION}
\turkcebaslik{SİMGELEME ÖĞRENEREK SİNİRSEL AĞLARLA İŞARET DİLİ ÇEVİRİSİ}
\degree{B.S., Computer Engineering, Bogazici University, 2017}
\author{Alptekin Orbay}
\program{Computer Engineering}
\subyear{2020}

\supervisor{Prof. Lale Akarun}
\examineri{Prof. Murat Saraçlar}
\examinerii{Assist. Prof. Furkan Kıraç}
\dateofapproval{18.08.2020}

\begin{document}

\pagenumbering{roman}
\makemstitle 
\makeapprovalpage
\begin{acknowledgements}
First, and most of all, I would like to thank my thesis advisor Prof. Lale Akarun
for her expertise, guidance, assistance and patience during my masters education. I
would also like to thank Prof. Murat Saraçlar and Assistant Prof. Furkan Kıraç for participating in
my thesis jury and their valuable comments and suggestions on my thesis.
\par
I would like to express my deepest gratitude to my family for their endless support
and encouragement throughout my life. Without them, I wouldn’t be the person who I
am today.
\par
I would like to thank my friends and colleagues in Boğaziçi University and Percep-
tual Intelligence Laboratory for their endless support and valuable friendship, namely,
Ahmet Alp Kındıroğlu, Do\u{g}a Siyli, Gizem Esra Ünlü,
Mehmet Burak Kurutmaz, O\u{g}ulcan Özdemir. I would also like to extend my thanks to Umut Başer, Ecem Alantuğ and Burcu Topçcu for relaxing coffee breaks in our busy schedules.

\par This thesis has been supported by the Scientific and Technological Research Council of Turkey (TUBITAK) Project \#117E059.

\end{acknowledgements}
\begin{abstract}
In this thesis, we propose a multitask learning based method to improve Neural Sign Language Translation (NSLT) consisting of two parts, a tokenization layer and Neural Machine Translation (NMT). The tokenization part focuses on how Sign Language (SL) videos should be represented to be fed into the other part. It has not been studied elaborately whereas NMT research has attracted several researchers contributing enormous advancements. Up to now, there are two main input tokenization levels, namely frame-level and gloss-level tokenization. Glosses are world-like intermediate presentation and unique to SLs. Therefore, we aim to develop a generic sign-level tokenization layer so that it is applicable to other domains without further effort.
\par We begin with investigating current tokenization approaches and explain their weaknesses with several experiments. To provide a solution, we adapt Transfer Learning, Multitask Learning and Unsupervised Domain Adaptation into this research to leverage additional supervision. We succeed in enabling knowledge transfer between SLs and improve translation quality by 5 points in BLEU-4 and 8 points in ROUGE scores. Secondly, we show the effects of body parts by extensive experiments in all the tokenization approaches. Apart from these, we adopt  3D-CNNs to improve efficiency in terms of time and space. Lastly, we discuss the advantages of sign-level tokenization over gloss-level tokenization. To sum up, our proposed method eliminates the need for gloss level annotation to obtain higher scores by providing additional supervision by utilizing weak supervision sources.
\vspace{-\baselineskip}
\end{abstract}
\begin{ozet}
	Bu tezde, Çoklu Görev Öğrenmesi tabanlı bir yöntem 
Sinirsel Ağlarla İşaret Dili Çevirisini geliştirmek için önerilmiştir. Sinirsel Ağlarla İşaret Dili Çevirisi, Sinirsel Ağlarla Makine Çevirisi ve Simgeleme olarak iki parçadan oluşmaktadır. Simgeleme tabakası ana odağına işaret dili videolarının nasıl bir gösterim ile diğer katmana beslenmesi gerektiğini ele alır. Simgeleme pek fazla keşfedilmiş bir alan değildir. Halbuki, Sinirsel Ağlarla Makine Çevirisi pek çok farklı disiplinden araştırmacının ilgisini çekmiştir ve bu sayede alanda sürekli ilerleme kaydedilmektedir. Simgeleme bugüne kadar çerçeve ve gloss olmak üzere iki seviyede incelenmiştir. Glosslar kelime düzeyinde anlam taşıyan ara gösterimlerdir ve her işaret diline özel anlam taşırlar. Ayrıca, gloss bazında etiketleme çok emekte gerektiren bir süreçtir. Bu yüzden, çerçeve seviyesinde Simgeleme ile herhangi bir işaret diline  kolay uygulanabilir yöntem geliştirmeyi hedefledik.  

	Öncelikle bugüne kadar sunulan yöntemlerin eksik yanlarını pek çok deney ile açığa çıkarttık. Elde ettiğimiz kazanımlar ile Öğrenme Aktarması, Çoklu Görev Öğrenme ve Gözetimsiz Alan Uyarlaması tekniklerini bu probleme uyarladık. Bu sayede etiketlenip anlamlandırılmış veri eksikliğini gidermeye çalıştık. İşaret dilleri arasında bilgi aktarımını sağlayarak BLUE-4 metriğinde 5 ve ROUGE skorunda 8 puanlık bir iyileştirme sağladık. İkinci olarak, insan vücut bölümlerinin anlama etkilerini her deneyimizde ayrıca inceledik. 3D-CNN yapısını zamansal ve mekansal bakımdan daha etkili bir sistem oluşturmak için uyguladık. Son olarak, işaret seviyesinde simgelemenin glossa göre avantajlarını açıkladık. Özetlemek gerekirse, sunduğumuz yöntem zayıf gözetim kaynaklarını kullanarak gloss bazında etiketlemeye olan bağlılığı azaltıyor.
\vspace{-\baselineskip}
\end{ozet}

\tableofcontents
\listoffigures
\listoftables
\begin{symbols}
%
\sym{$a_t$}{Attention weights at step $t$}
\sym{$c_t$}{Context vector at step $t$}
\sym{$e_{1:Z}$}{Distance scores of all encoder outputs}
\sym{$f$}{Encoder Function}
\sym{$g$}{Decoder Function}
\sym{$f \circ g$}{Decoder Function}
\sym{$h_{1:S}$}{Decoder outputs from step 1 to $S$}
\sym{$h_f$}{Encoder hidden states}
\sym{$L$}{Loss Term}
\sym{$o_{1:Z}$}{Encoder outputs from step 1 to step $Z$}
\sym{$t_{1:Z}$}{Tokens from step 1 to step $Z$}
\sym{$T_x$}{Normalized x coordinate values of hand keypoints}
\sym{$T_y$}{Normalized y coordinate calues of hand keypoints}
\sym{$V_x$}{x coordinate values of hand keypoints}
\sym{$V_y$}{y coordinate values of hand keypoints}
\sym{$w_{1:S}$}{Target words from step 1 to step $S$}
\sym{$x_{1:N}$}{Frames from step 1 to step $N$}

\sym{}{}
\sym{$\alpha$}{Distance function between output vectors}
\sym{$\beta(t)$}{Loss term weight at time step t}
\sym{$\theta$}{Dense layer}
\sym{$\lambda(t)$}{Coefficient in Gradient Reversal Layer at iteration t}
\sym{$\sigma$}{Standard Deviation}
\sym{$\psi$}{Tokenization Function}

\end{symbols}

\begin{abbreviations}
\sym{1D}{One Dimensional}
\sym{2D}{Two Dimensional}
\sym{3D}{Three Dimensional}
\sym{2D CNN}{Two Dimensional Convolutional Neural Network}
\sym{3D CNN}{Three Dimensional Convolutional Neural Network}
\sym{ASL}{American Sign Language}
\sym{BSL}{British Sign Language}
\sym{CSL}{Chinese Sign Language}
\sym{CSLR}{Continuous Sign Language Recognition}
\sym{CNN}{Convolutional Neural Network}
\sym{CTC}{Connectionist TemporalCclassification}
\sym{ED}{Encoder-Decoder}
\sym{DTW}{Dynamical Time Warping}
\sym{HOF}{Histogram of Oriented Gradients}
\sym{HOF}{Histogram of Optical Flow}
\sym{HMM}{Hidden Markov Model}
\sym{GSL}{German Sign Language}
\sym{GRL}{Gradient Reversal Layer}
\sym{GRU}{Gated Recurrent Unit}
\sym{IDT}{Improved Dense Trajectories}
\sym{ILSVRC}{ImageNet Large-Scale Visual Recognition Challenge}
\sym{IOHMM}{Input-Output Hidden Markov Model}
\sym{LSTM}{Long-short Term Memory}
\sym{NMT}{Neural Machine Translation}
\sym{NSLT}{Neural Sign Language Translation}
\sym{PH}{RWTHPHOENIX-Weather 2014 }
\sym{RNN}{Recurrent Neural Network}
\sym{SL}{Sign Language}
\sym{SLR}{Sign Language Recognition}
\sym{TSL}{Turkish Sign Language}
\end{abbreviations}

\chapter{INTRODUCTION}
\label{chapter:introduction} 
\pagenumbering{arabic}

Sign Languages (SL) are the main way of communication for hearing impaired people. Deaf people in a region develop a unique sign language independent of the region's spoken languages. In addition to that, there is no strong relation between different sign languages. 
Therefore, SLs are considered as natural languages like spoken and written languages. This means SLs have their own linguistic properties as well as visual properties. Hence, it is required to identify their complex structures and treat them as separate languages.

\par
\begin{figure}[htbp]
\begin{center}
\includegraphics[width=1\columnwidth]{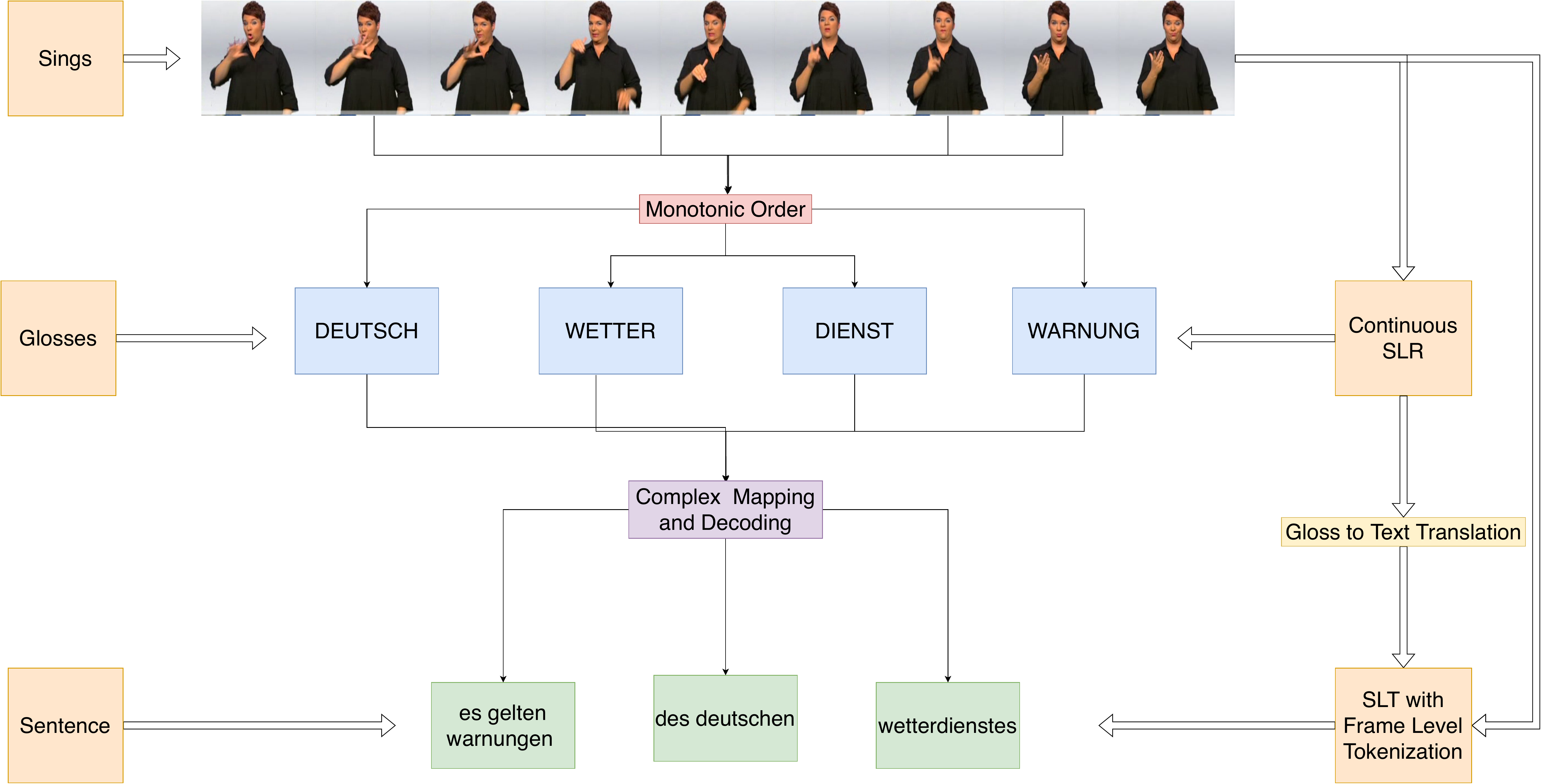}
\end{center}
\caption{Important elements and steps in Sign Language research.}
\vskip\baselineskip
\label{fig:slt-slr}
\end{figure}

\begin{figure}[htbp]
\begin{center}
\includegraphics[width=1\columnwidth]{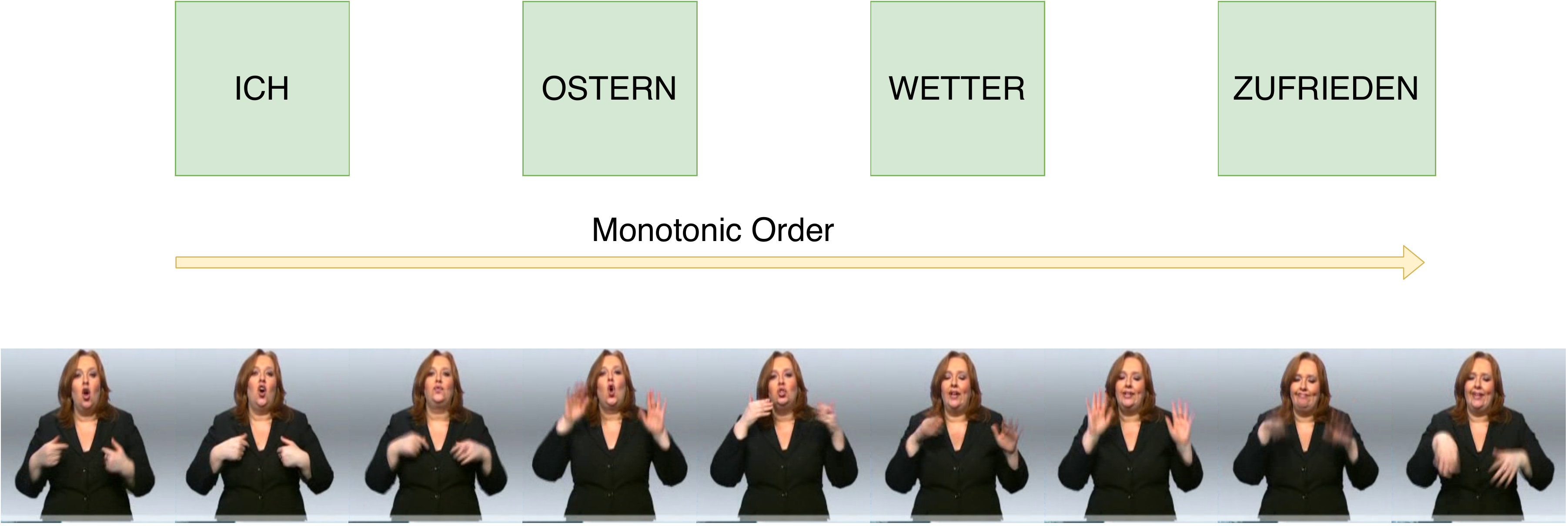}
\end{center}
\caption{An illustration of glosses and signs in a German Sign Language sentence.}
\vskip\baselineskip
\label{fig:slr-illus}
\end{figure}

 There have been several studies to bridge the communication gap between spoken and sign languages. Different concepts have been developed to analyze SLs in detail as seen in Figure \ref{fig:slt-slr}. First of all, signs are semantic parts in SL videos. In a SL video, all frames are usually not meaningful. In general, some frames are transition frames whereas some are keyframes where signers perform special hand shapes. The other key concept is gloss. Glosses are word-like representations corresponding to signs in terms of meaning. Glosses are in text form, but they do not construct a full sentence if they are ordered in the same manner as signs. As seen in Figure \ref{fig:slr-illus}, glosses in German Sign Language (GSL) are illustrated from the example frame sequence. "ICH" means "I" in English. In this example, it means "we", but corresponding signs may have different meanings like "my", "I", "me" in a different video. This approach generates meaningful interpretations; however, linguistic properties are not ignored. Furthermore, the hand shapes in SL videos require special attention as they do not have the same meanings as daily life gestures. Hence, hand shapes should be studied very carefully. Another challenge is that subtle changes in SL hand shapes may lead to dramatic differences in meaning. 

\par Sign Language Recognition refers to a number of recognition tasks. In the isolated setting, it may refer to identifying the labels of a sign; which are often glosses. In the continuous setting, it refers to a broad range of tasks, from sign spotting to sign sentence identification. As stated earlier, hand shape recognition is crucial, but hand shapes differ from one sign language to another semantically and syntactically. In addition to that, the conveyed information is not limited to hand shapes, but also general frame contexts play an important role. The second challenge is that the meaning of signs is dependent on the overall context. This is to say that a proper approach should cover the relationship between different signs despite the fact that the order of glosses and signs is monotonic in a sentence. In short, a good approach should be holistic to attain successful results for SLR. 

\par
\begin{figure}[htbp]
\begin{center}
\includegraphics[width=1\columnwidth]{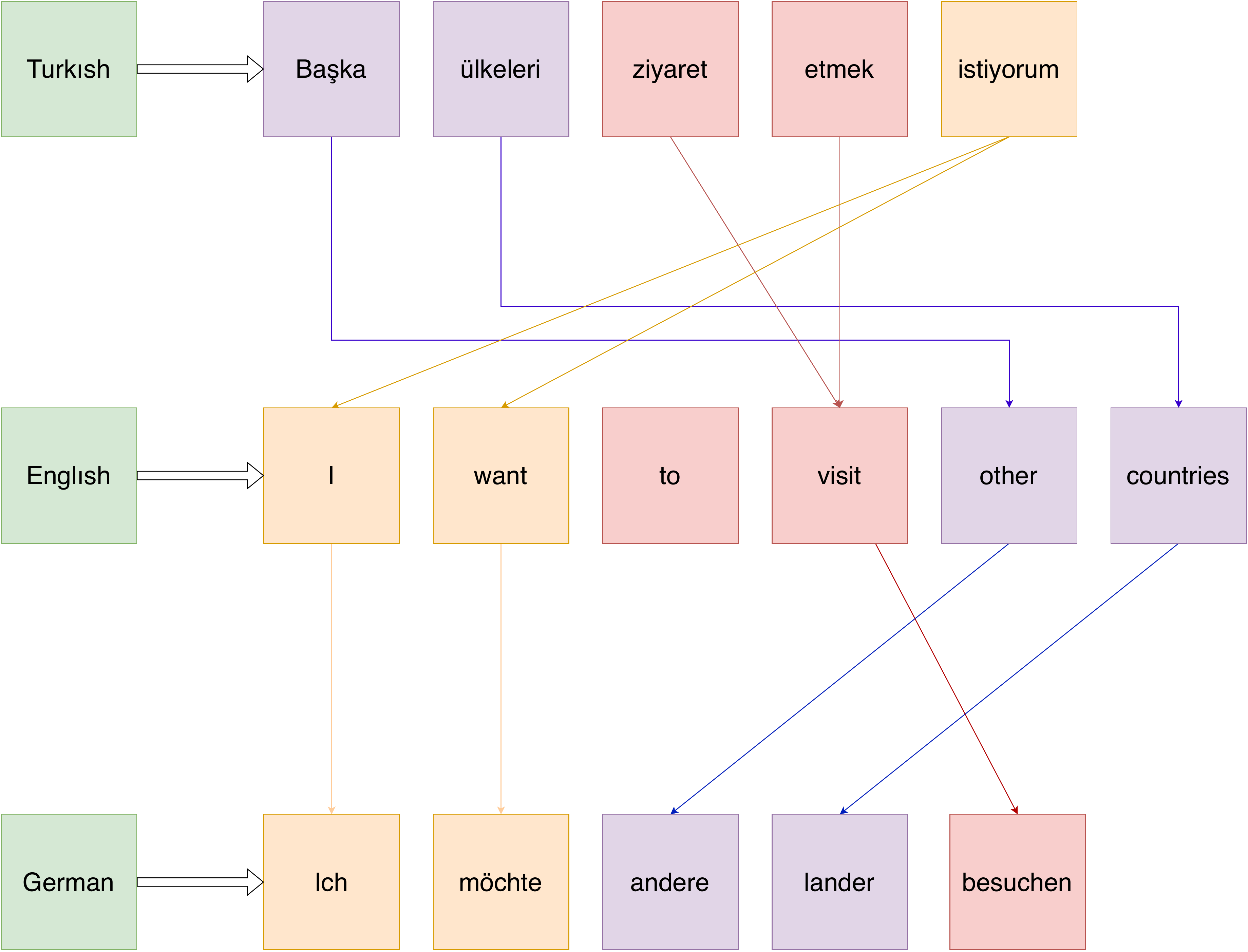}
\end{center}
\caption{An example of translation between three different spoken languages.}
\vskip\baselineskip
\label{fig:spoken-trans}
\end{figure}

\begin{figure}[htbp]
\begin{center}
\includegraphics[width=1\columnwidth]{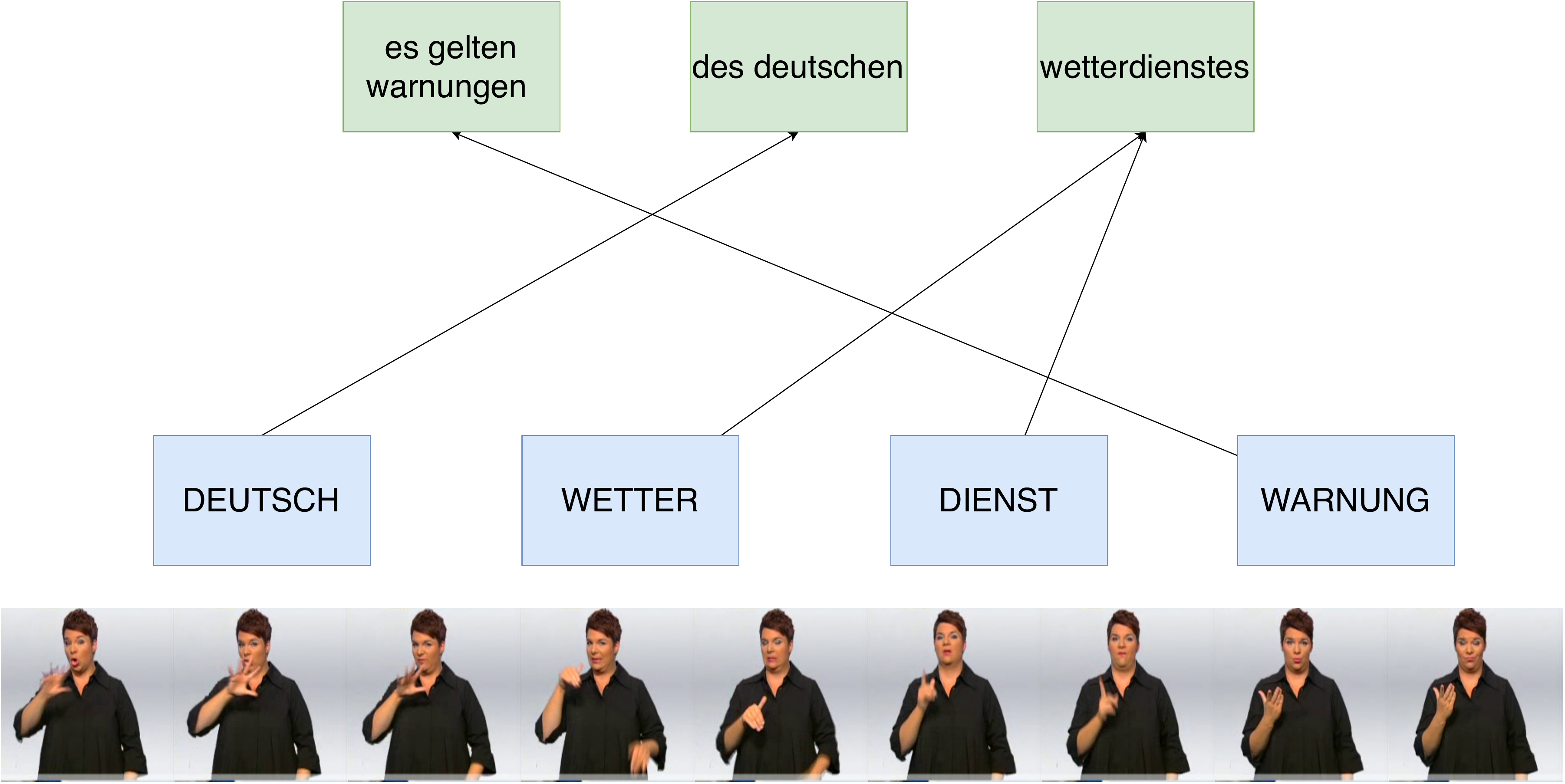}
\end{center}
\caption{An example of Sign Language translation.}
\vskip\baselineskip
\label{fig:slt-example}
\end{figure}

 Recently, another research field, Neural Sign Language Translation (NSLT), has emerged to provide translations from sign language to spoken languages just like translation from a spoken language to another.
Sign Language Translation (SLT) shares the same basics with other translation tasks as seen as seen in Figure \ref{fig:spoken-trans} and  Figure \ref{fig:slt-example}. In addition to that, SLT requires further effort to tackle with its visual properties. An example translation between three different languages is shown in Figure \ref{fig:spoken-trans}. The three sentences have the same meaning, but there are certain differences such as the lengths of sentences and the order of the words. Therefore, word by word translation is not the proper approach. Furthermore, word meaning is dependent on the whole sentence. For example, "to" in an English sentence may have several meanings. We can choose the correct one in this sentence by examining the words placed before and after it. Another important point is that the output words must not violate the grammar of the target language. To conclude, the solution should cover all the concerns and those rules are also prevailing for SLT.

\par NSLT aims to enable continuous dialog between hearing-impaired individuals and people who do not know SLs. Neural Machine Translation (NMT) has achieved several advancements to sustain human-level translations. Therefore, NSLT may use the same methods and findings, but the visual properties require new approaches. NMT formulation may be adapted by NSLT. Firstly, the source and target sentences are represented with tokens. Tokens are considered as atomic and meaningful blocks in sentences. Tokenization is the name of the process to find the most appropriate tokens for the translation. Splitting words in a sentence is the most straightforward and effective approach. However, agglutinative languages are not suitable for word-level tokenization so n-gram chars are widely used for this type of languages. This implies that tokenization strategy must be tailored for each target domain. After choosing a proper tokenization method, we can continue to formulate the problem as follows. There are $(x_1, x_2, \dots , x_n)$  symbols as input tokens, $\bold x$. The symbols and their order represent a context, $c_{\bold x}$. The target symbols are $(y_1, y_2, \dots , y_n)$ as output tokens, $\bold y$. Eventually, a complex function $f((x_1, x_2, \dots , x_n),c_{\bold x}) \rightarrow (y_1, y_2, \dots , y_n) $ should be found. NMT researchers have proposed several architectures and methods to find the most optimal mapping between the tokens. However, in SLT research, we need a special tokenization approach to apply NMT architectures successfully.

\section{Motivation}
	In this section, we mention different tokenization techniques for SLT and explain our perspective on the problem. We mentioned about the basics of SLT and NMT. From our research perspective, NMT methods can provide successful results if we have good tokens from SL vides. Therefore, tokenization is seen as the most crucial part of this research. Firstly, the visual properties are involved in the tokenization part. Secondly, there is no a generic approach to obtain strong tokens for all the SLs. In addition to that, it is not clear that discrete tokens should be obtained for better translation quality. For this reason, we extend the meaning of tokenization for NSLT and it covers the overall process to prepare the frames for the NMT module.
	\par
\begin{figure}[htbp]
\begin{center}
\includegraphics[width=1\columnwidth]{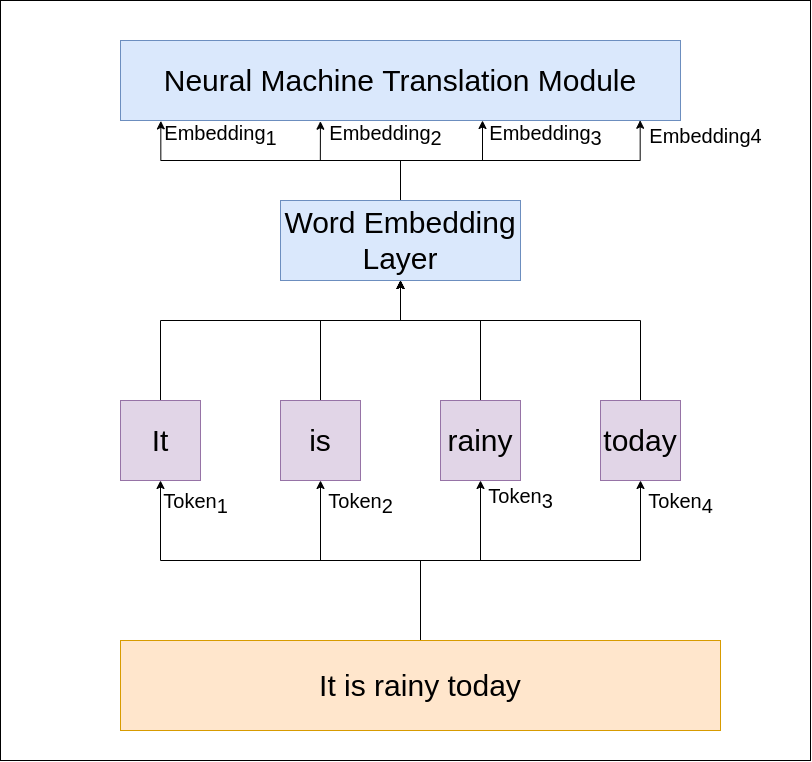}
\end{center}
\caption{An illustration of word-level tokenization in NMT.}
\vskip\baselineskip
\label{fig:nmt-token}
\end{figure}

 For spoken to spoken languages, we generally use words as tokens to feed the NMT module. The current state-of-the-art method converts those tokens to continuous embeddings to reach a semantic representation. While learning translation, the word embedding is also trained to learn the relationship between words. Eventually, a meaningful embedding is obtained before the NMT module as seen in Figure \ref{fig:nmt-token}. Based on this, it may be a good idea to learn a good representation of signs to replace with word embeddings to achieve the same advancements in NSLT as NMT has done. This representation is cross-lingual; but learning it is an open problem. Our research is mainly focused on this problem. Before introducing our approach, we discuss the existing three tokenization approaches in the following subsection.

\subsection{Input Tokenization in NSLT}

\par  The first approach is using glosses as tokens. Glosses are intermediate word-like representations between signs and words in sentences. Therefore, they can be directly applicable to the NMT framework without any further effort. However, there are certain shortcomings in this method. Firstly, glosses rarely exist in real life. Gloss annotation requires a laborious process and special expertise. Secondly, glosses are unique to SLs. Therefore, each SL requires special effort to obtain glosses whereas sentences are commonly available. The last drawback is that a mistake in the gloss level can produce dramatic meaning differences in translation, since glosses are high level annotations, similar to words.

\par
\begin{figure}[htbp]
\begin{center}
\includegraphics[width=1\columnwidth]{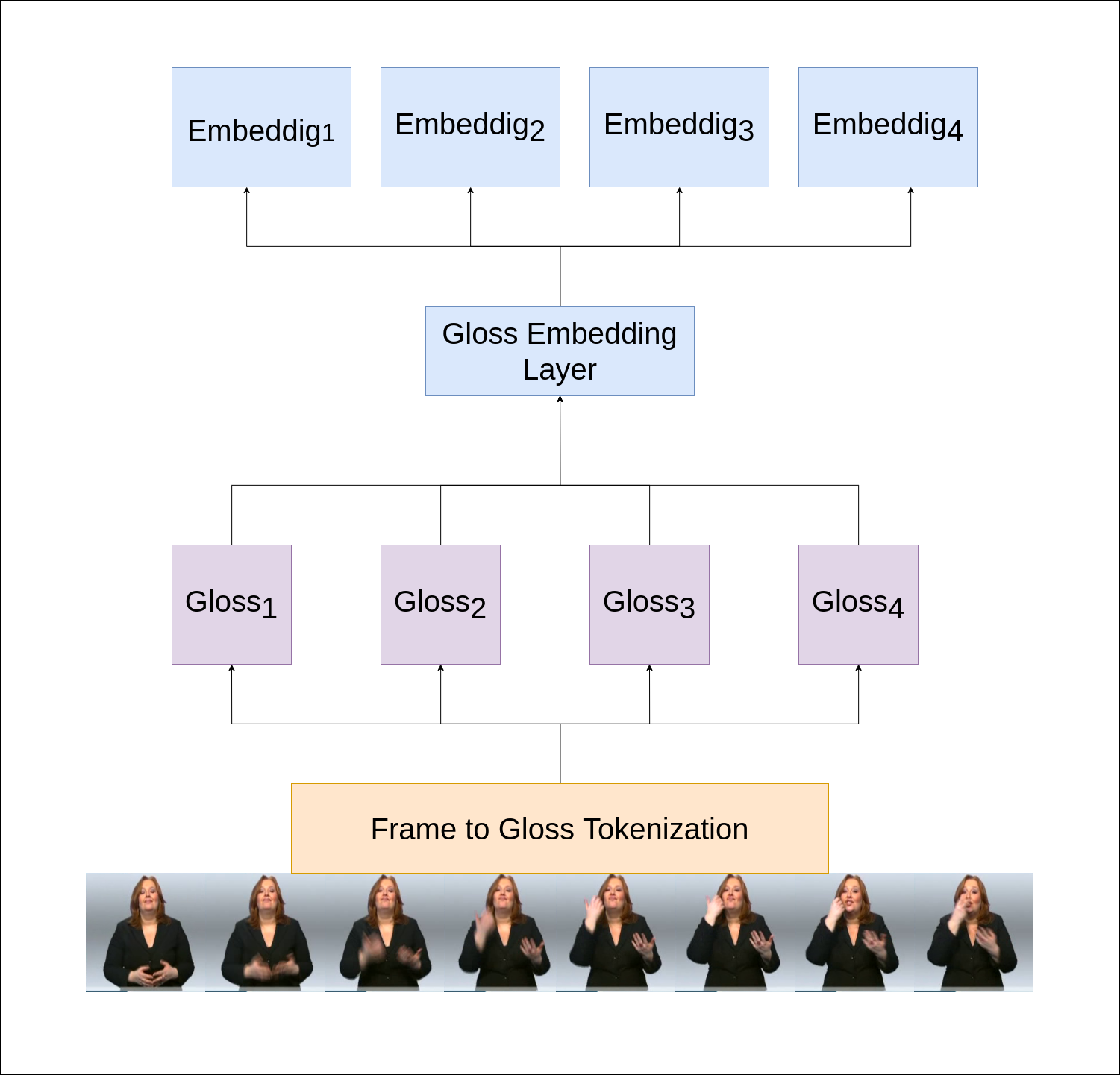}
\end{center}
\caption{An illustration of frame to gloss tokenization in NMT.}
\vskip\baselineskip
\label{fig:frame-gloss}
\end{figure}

 The second approach is the same as the first one in terms of tokens. On top of that, this approach learns to extract glosses from frames. In other words, this method uses glosses as explicit intermediate representations as seen in Figure \ref{fig:frame-gloss}. It eliminates the further search for tokenization, but it needs a special network for frame to gloss conversion. There are two main concerns. The first one is that a network for frame to gloss conversion is still dependent on gloss annotations. The second is that it is not clear that glosses are the upper bound for SLT as there is not sufficient evidence. The problem is immature and the result in \cite{Camgz2020SignLT} provides clues about whether glosses may restrict translation quality.

\begin{figure}[htbp]
\begin{center}
\includegraphics[width=1\columnwidth]{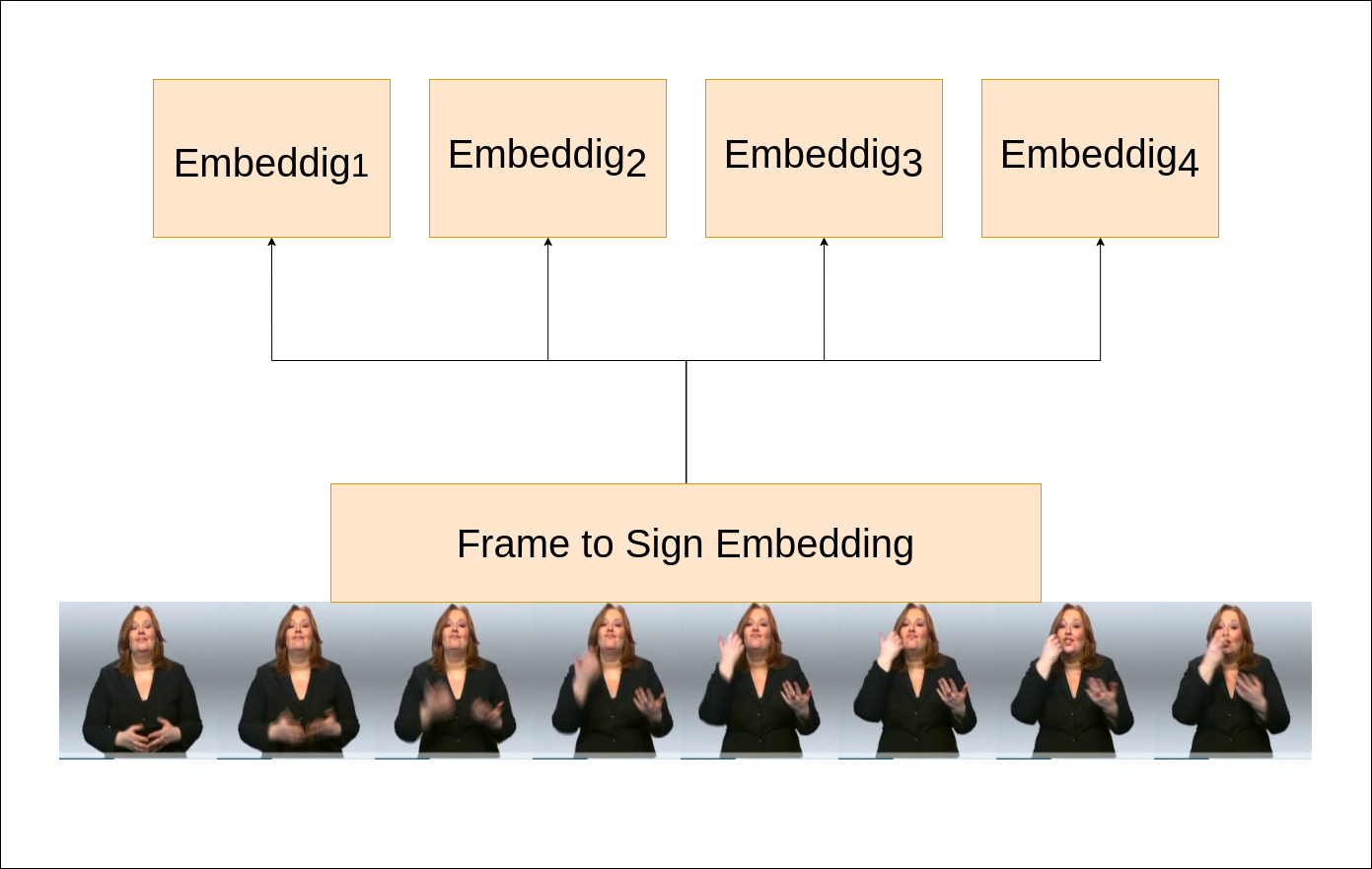}
\end{center}
\caption{An illustration of frame level tokenization in NMT.}
\vskip\baselineskip
\label{fig:frame_level}
\end{figure}

The third approach is called frame-level tokenization. This approach does not establish any explicit intermediate representation as seen in Figure \ref{fig:frame_level}. It aims to learn good sign embeddings to replace with word-embeddings. However, there is no golden way to represent signs with embeddings to feed into the NMT module. Furthermore, it is not clear what the length of the embedding should be. Embeddings can be obtained from each frame or extracted from inner short clips in the video. In addition to that, the representation can be learned with sentence-video pairs or trained outside the NSLT system. There are several ways for frame-level tokenization. However, the main difference from the gloss level tokenization is that discrete representation can be eliminated. If we find a proper one, there would be several advantages. The first one is that the resulting framework can be applied to any SL translation task without requiring annotation. The second advantage is the opportunity to inject additional supervision. The representations would be trained on different tasks and different datasets whereas gloss level tokenization cannot cover different SLs. The third one is that the token length can be adjusted. To boost translation speed, the number of tokens can be reduced to a pre-determined number.

\section{Research Overview and Contributions}
\label{list:cont}
The study focused mainly on searching alternative tokenization methods. We have three main motivations. The first one to leverage additional supervision to SLT. We suspect that the data amount is limited. Generally, SL data is very scarce and we would like to benefit from all sorts of data with maximum efficiency. The second one is that our proposed approach would enable a generic tokenization layer. In this manner, the method can be adopted by any SL. The third one is providing flexibility for real-time applications. The current approaches require a vast amount of computational power. Therefore, we search for ways to increase efficiency and analyze the effects on translation quality. 
	
\par Our approach is well suited to benefit from different datasets and different sign languages to leverage additional supervision. We claim that general purpose datasets are sufficient to provide better translations. As a result, the need for annotation from domain experts is eliminated. In this manner, the first and second are addressed. We also show a trade-off between time efficiency and translation quality. The overall contributions of this study can be summarized as:
	\begin{itemize}
    \item It illustrates that end-to-end learning is improved by learning tokenization explicitly. 
    \item It proposes a semi-supervised tokenization method utilizing Multitask Learning, Domain Adaptation and Transfer Learning to leverage additional supervision.
    \item  It shows the effectiveness of 3D-CNN in Sign Language Translation.
  \item It proves the effectiveness of the proposed method by comparing other tokenization methods trained on target domain data. 
  	\end{itemize}

  \par	First, we investigate end-to-end learning scheme proposed by Camgoz et al. \cite{camgoz2018neural} which formalized Neural Sign Language Translation (NSLT) for the first time. The formalized system mainly consists of two parts, tokenization and NMT. This study reveals the shortcomings of the newly proposed NSLT system under different input settings. The results of the initial experiments indicate that it has two drawbacks. The first one is that it cannot attach enough attention to the body parts. The data amount in training is not sufficient to make the model fully functional.
 \par To deal with the first drawback, this study addresses the tokenization part of the system. Along with the tokenization approach with human keypoint proposed by Ko et al. \cite{nslt-humankeypoint}, it conducts several experiments with new tokenization methods. As a result, it illustrates the effectiveness of alternative tokenization methods over the end-to-end learning scheme.
 \par It is known that annotation in SL requires additional effort and expertise, Hence, the study utilizes Multitask Learning, Domain Adaptation and Transfer Learning methods to develop an efficient tokenization approach in the absence of insufficient amount of data. In other words, the main contribution of this study is that it proposes a state-of-the-art tokenization approach that is independent of the target domain.
 \par Moreover, the study tests the proposed method in different experimental setups. As the data amount for translation is very limited and NMT models are large enough to overfit training datasets easily, it should be shown that the tokenization part is generic to function in other SL datasets. Therefore, the study divides experiments into two parts. One is tokenization learning involving target domain data and the other is without target domain knowledge. Eventually, the study concludes that the proposed method can close the gap between frame level and gloss level tokenization approaches even if target Sign Languages are not seen in learning tokenization.
 
\section{Thesis Outline}
\label{list:outline}

The rest of the thesis is organized as follows: In Chapter 2, we review the literature
on both sign language recognition and sign language translation. We also investigate the popular datasets
for both domains in the same chapter. In Chapter 3, we explain the background of NSLT, Multitask Learning and Domain Adaptation.
 After we explain their background information, we introduce our experimental setups and explain the usage of the proposed methods along with other alternatives in the literature in Chapter 4. The chapter also presents experiment results and their implications. Lastly, we share our final thoughts about our work and give some future
directions in Chapter 5.

\chapter{RELATED WORK}
\label{chapter:related-work}

\section{Sign Language Recognition}
\label{list:slr}
	Sign language research is a special area that requires a multi-disciplinary approach to conduct detailed studies. The development of this research is highly dependent on data collection as its annotation process is very laborious and requires special knowledge. However, there are some dictionaries \cite{danish}, \cite{nz}, \cite{tid} for educational purposes. These types of sources are not useful unless they include an automatic labeling strategy. 

\par Some attempts are performed to collect data with wearable sensors. Those provide very informative measurements like angles between fingers, coordinates of key points and velocities. In \cite{power-glove}, PowerGlove is used to collect data and hand-crafted features from those measurements are fed to decision trees. Magnetic sensors are used to collect American Sign Language (ASL) data in \cite{hmm-asl}. These approaches are important to reveal dynamics of Sign Languages, but not applicable in the wild. 

\par As a result, researchers have begun to 
collect datasets \cite{chinesse}, \cite{bosphorus}, \cite{ebling-etal-2018-smile} in controlled environment with RGB cameras. They are beneficial for identifying signs but they are very limited in terms of variation and context. They have paved the way for developing bigger datasets with their supervision.

\par Researchers have continued to search for ways to construct larger datasets without additional efforts. The studies in \cite{Pfister}, \cite{Buehler} ,\cite{hellen} have utilized weak labels in public broadcasts for hearing impaired people. This type of data can be aligned with subtitles or lip reading. Thanks to a huge amount of data, those approaches have been accomplished to create big datasets by utilizing those side information. Forster et al. released RWTH-PHOENIX-Weather 2012 \cite{forster1} and later its extended version RWTH-PHOENIX-Weather 2014 \cite{forster2} from TV broadcasts for hearing impaired people. Eventually, a dataset is publicly available to researchers with continuous gloss and sentence annotations.

\par In previous studies, hand-crafted features are used to represent signs in images. Dalal \etal \cite{hog} proposed Histogram of Oriented Gradients (HOG) to detect humans in images. Liwicki \etal \cite{hog-sl} modifies this hand shape descriptors to apply on British Sign Language (BSL) alphabets. Lim \etal \cite{hof-sl} uses Histogram of Optical Flow (HOF)  \cite{hof}. Ozdemir \etal \cite{idt-sl} applies Improved Dense Trajectories (IDT) \cite{idt} on Isolated Turkish Sign Language (TSL) data. They also extend their work in \cite{idt-sl-fast} as IDT is computationally expensive. HOG is calculated on blocks in images with orientation gradients to represent appearances and shapes of objects. HOF is similar to HOG, but extracted from optical flows in consecutive frames. Those descriptors are very robust to different environmental conditions.

\par Dynamical Time Warping (DTW) \cite{dtw} is a technique that finds patterns in sequences with dynamic programming. Hidden Markov Models (HMM)\cite{hmm} is a graph-based probabilistic for temporal modeling. HMMs are generative models, but Bengio \etal \cite{iohmm} proposed Input-Output HMM which has discriminant properties. Vogler \etal \cite{hmm-asl} adapts HMMs to recognize ASL. Note that every sign must be modeled with a different HMM so pruning is needed to classify with lots of unique signs. Keskin \etal \cite{keskin} uses DTW for unsupervised clustering of sequences and IOHMMs to predict hand gestures on the clustered data. DTW generates a similarity score between sequences and IOHMM is learned to identify signs by their discriminative power.  

\par While the prior techniques boost sign language research in terms of data and domain knowledge, Deep Learning (DL) opened a new era for Computer Vision research. Krizhevsky \etal \cite{alexnet} made a breakthrough in image recognition on Imagenet-LSVRC contest. The success of their approach is based on the advanced capabilities of CNNs to represent images compared to hand-crafted features. Moreover, sequence modeling with Long Short Term Memory (LSTM) \cite{lstm} has become one of the most important advances in DL. Several achievements are accomplished in speech recognition \cite{pmlr-v48-amodei16}, speech translation \cite{las} and image captioning \cite{sat}.

\par With new datasets and DL techniques, Continuous Sign Language Recognition (CSLR) research has been boosted dramatically. CSLR maps input frames to glosses while SLR generally deals with isolated datasets. Koller \etal \cite{deephand} uses CNNs as frame-level feature extractor and HMMs so as to model of transition between frames. The models are trained jointly with Expectation-Maximization method. The study accomplishes to train on over one million images. Another important result is that the features extracted from the CNN outperform hand-crafted features (HOG-3D) dramatically in CLSR. In the study of \cite{deep-sign}, CNN-HMM architectures are directly trained on continuous gloss recognition. HMMs maximize the probability of gloss alignment and CNNs learn representations of images to construct hidden states of HMMs. Later, Bidirectional LSTMs (bi-LSTM) \cite{bi-lstm} are deployed between CNNs and HMMs to encapsulate contextual information in \cite{re-sign}. LSTM outputs are fed as hidden states of HMMs. The emergence of Connectionist Temporal Classification (CTC) proposed by Graves \etal \cite{CTC} enables optimize deep architectures in an end-to-end manner. CTC is a loss layer that calculates a penalty between two sequences following a monotonic order. While previous studies use Bayesian optimization with HMMs, Camgoz \etal replace HMMs with LSTMs for continuous hand shape recognition and outperforms DeepHand (CNN-HMM) \cite{deephand} on weakly annotated datasets.

\par Moreover, Carreira \etal \cite{I3D} achieves successful results on Kinetics dataset with 3D-CNNs architectures. Compared to CNN-LSTM models, 3D-CNNs can produce spatio-temporal embeddings with additional convolution operations in time. 3D-CNNs are adapted to gesture recognition tasks in \cite{3d-cnn-gesture-1}, \cite{3d-cnn-gesture-2}. Yang1 \etal \cite{sfnet} propose a architecture utilizing 2D and 3D convolutions to improve image representation quality for SLR.

\section{Sign Language Recognition Datasets}
\label{list:slr-dataset}
	We are interested in the two datasets. The first one includes over one million images proposed by Koller \etal \cite{deephand}. Its labeling strategy is determined by linguistic research. They claim that the proposed hand shapes cover all of the possible signs in different sign languages. The dataset consists of three different sign languages, namely German (ph), New Zealand (nz) and Danish SL. The major challenge in this dataset is that labeling involves a high level of noise. 
	\par The German dataset consists of the training set of RWTH-PHOENIX-Weather 2014 \cite{forster2}. Danish \cite{danish} and New Zealand \cite{nz} datasets are fetched from online dictionaries. Those are labeled for 61 classes with one junk class. As seen in Figure \ref{table:one-million-images}, RWTH-PHOENIX-Weather 2014 is so larger than others and involves many more signs and variations. Note that most of the frames are automatically labeled as junk class. 
	
\begin{table}[thbp] 
\caption{Corpus statistics of hand shape datasets.}
\label{table:one-million-images}
\begin{center}
\begin{tabular}{|c|c|c|c|c|}
\hline
 
  & danish & nz & ph & TSL \\ \hline \hline
frames              &145,720 & 288,593 &  799,006 & 56110\\ \hline
hand shape frame    & 65,088 & 153,298 & 786,750 & 56110                   \\ \hline
garbage frames      & 80,632 & 135,295 & 12,256 & 0                   \\ \hline
signs              & 2,149 & 4,155  & 65,227 & 30                     \\ \hline
signers               & 6  & 8  & 9 & 7                     \\ \hline
unqiue hand shapes  & 61 & 61 & 61 & 45         \\ \hline
\end{tabular}
\end{center}
\end{table} 

\begin{figure}[htbp]
\begin{center}
\includegraphics[width=1\columnwidth]{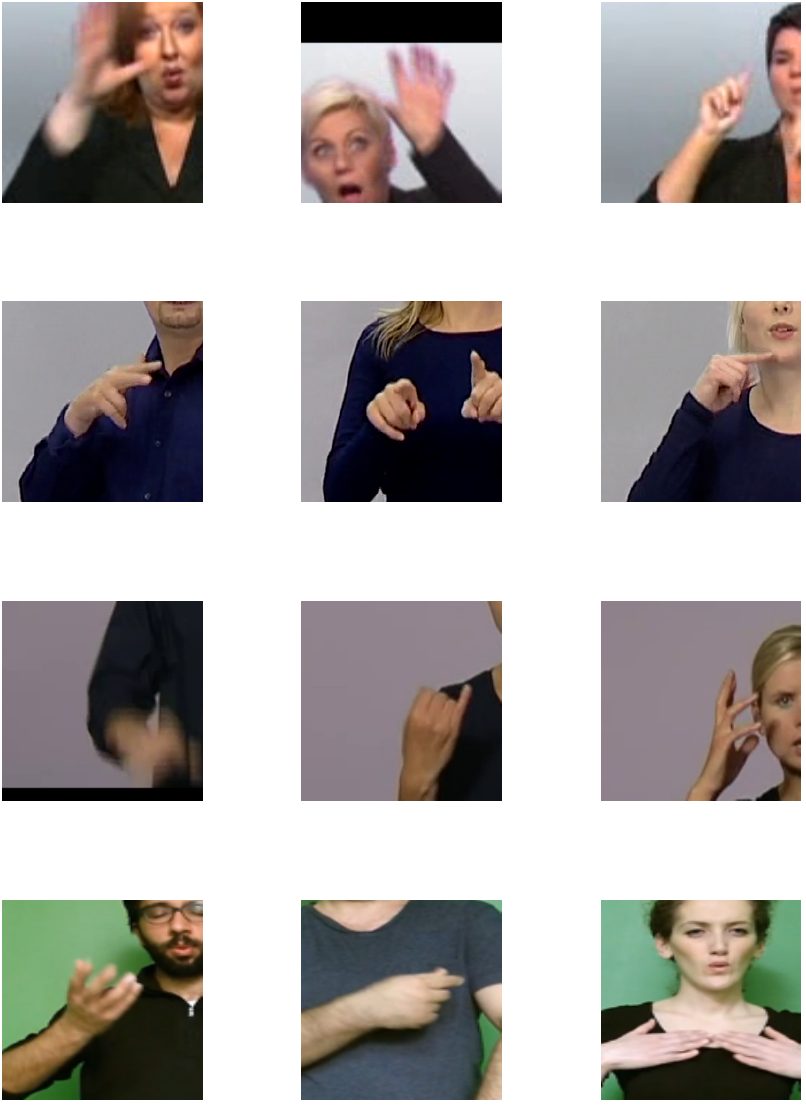}
\end{center}
\caption{Three sample images from each  data source. Top to bottom ph, danish, nz, TSL.}
\vskip\baselineskip
\label{fig:sample-img}
\end{figure}

	The second main dataset is Turkish Sign Language (TSL) introduced by Siyli \etal \cite{doga}. This dataset is too small compared to other datasets. However, the dataset is labeled in an unsupervised manner and hand shapes are determined in a data-driven fashion instead of linguistic research. Moreover, this dataset labels convey semantic meaning and more information compared to others. In other words, labels do not only mention hand shape, but also interactions of hands with other body parts. Some samples from the four datasets are seen in Figure \ref{fig:sample-img}.

\begin{sidewaystable}[thbp]
\thisfloatpagestyle{empty}
\vskip\baselineskip 
\caption{Datasets used in the literature of Sign Language Recognition.}
\begin{center}
\resizebox{\textwidth}{!}{
\begin{tabular}{|l||c|c|c|c|c|c|}
 \hline
Dataset Name & Year &  Signs  & Samples &  Users & Modality & Environment \\
  \hline
SIGNUM \cite{signum}& 2010 &  450   &  33,210 utterances &  25 & RGB & Controlled \\
  \hline
 RWTH-PHOENIX-Weather 2014 \cite{forster2}& 2014 &  1231  & 6931 Sentences  &  9 & RGB & Wild \\
  \hline
DEVISIGN \cite{devisign} & 2015 &  2000   & 24000 &  8 & RGB-D  & Controlled \\
  \hline
ChaLearn LAP Gesture \cite{chalearn} & 2016 &  249   & 47933 &  21 & RGB-D  & Controlled \\
  \hline
BosphorusSign \cite{bosphorus}  & 2016 &  636    & 24161 &  6 & RGB-D & Controlled \\
  \hline
  Video-Based CSL \cite{video-based-csl} & 2018  &  178    & 25000 &  50 & RGB & Controlled \\
  \hline
  MS-ASL  2019 \cite{ms-asl} & 2019 &  1000    & 25513 &  222 & RGB & Wild \\
  \hline
\end{tabular}}
\label{table:slr-database}
\end{center}
\end{sidewaystable}

Apart from these, there are other SLR datasets collected with different methods. The datasets are listed in Table \ref{table:slr-database}. SIGNUM \cite{signum} is a German Sign Language Dataset involving isolated and continuous videos. It covers 450 signs and 780 sentences. DEVISIGN is a Chinese Sign Language (CSL) dataset consisting of 2000 isolated signs in 24000 videos. In the work of \cite{video-based-csl}, there are 100 continuous sentences which are performed 5 times with 50 signers. This is also collected in a laboratory environment. ChaLearn LAP Gesture dataset \cite{chalearn} is a general gesture recognition dataset. There are 259 gestures performed by 21 different individuals. BosphorusSign \cite{bosphorus} is collected with Microsoft Kinect v2 depth sensor that provides RGB along with depth information. It includes signs from health, finances and most common expressions in Turkish. The dataset covers 636 isolated signs with 24161 videos. MS-ASL \cite{ms-asl} is a American Sign Language (ASL) dataset. The labeling is done at gloss-level. The videos are captured in several background conditions and there are 222 unique signers are involved in 25513 videos.

\section{Neural Machine Translation}
\label{lit:nslt}
	Neural Machine Translation aims to generate a meaningful sentence in target languages when a sentence in source languages is given. It learns the linguistic properties of source and target languages, word similarities at the same time. Mikolov \etal \cite{word2vec} shows the effectiveness of word embeddings in an unsupervised manner. So, NMT models have begun to adapt word embeddings to obtain more informative word representations for the source and target instead of one-hot vectors. However, prediction whole sequences from the beginning sometimes are cumbersome. The reason is that initial predictions of target words are generally noisy. This harms the learning procedure as the models generate word sequences depending on previously predicted words. Bengio \etal \cite{teacher-forcing} proposes a procedure that feeds golden labels as predictions to stabilize the learning in the first steps of learning. 
	\par It is a very challenging problem to evaluate human translations. In fact, there may be several perfect translations for a given sentence. Also, spoken language grammars allow sentences to be in different forms with the same words. It is more complicated to evaluate machine translations as they are applied to huge datasets consisting of millions of sentences. There are three popular evaluation metrics for automatic language translation in NMT. The first one is METEOR proposed by Banerjee \etal \cite{meteor}. The formulation is as follows:
\begin{align}
	P &= \frac{m}{w_t} \CommaPunct \\
	R &= \frac{m}{w_r} \CommaPunct \\
	F &= \frac{10PR}{R + 9P} \CommaPunct  \\
	p &= 0.5 (c/u_m)^3 ,   \\
	M &= F(1-p) ,
\end{align}
where $m_t$ is the number of matched unigrams between candidate and reference sentences. $w_t$ and $w_r$ are the number of unigrams in candidate and reference sentences. $c$ is the fewest possible unigram chunks adjacent in the reference and candidate. $u_m$ is the number of unigrams in the matched chunks. $M$ score converges 1 if the perfect matching occurs with an infinite number of words. For a perfect match with a three-word sentence, $c$ is 1  and $u_m$ is 3 where $M$ is 0.96. The second method is Recall-Oriented Understudy for Gisting Evaluation (ROUGE) \cite{rouge}. ROUGE-n means the ratio of overlapped n-grams over total n-grams between reference and candidate translations. The most popular one is Bilingual Evaluation Understudy (BLEU) proposed by Papineni \etal \cite{bleu}. BLEU modifies precision with clipped counts compared to the equations above. If an n-gram is matched several times with n-grams in references, the number of matches is restricted with the maximum number of counts in the references. For example, "a" is a very common stop word in English. If a candidate sentence consists of only this stop word, every n-gram in a candidate can be matched. Therefore, it reduces the score of junk sentences. Moreover, BLEU-n calculates the precision of all n-grams in the candidates and takes a geometric average of them. To enforce a penalty to shorter candidates than references, it multiplies the geometric mean with $\exp(1 -r/c)$ where $r$ is the reference length and $c$ is the candidate length. However, those attempts are not perfect to evaluate translations for two reasons. First of all, every word has many synonyms that implies the same meanings. Secondly, these approaches do not have a holistic grammatical perspective. A candidate sentence can have a high score in the metrics but does not have a meaning. Hence, two or more distinct metrics are indicated to be more confident about the quality.

\par While language models are improved by Natural Language Processing (NLP) methods, DL made a huge impact on sequence-to-sequence models. After LSTMs are shown to be effective for language modeling in \cite{lstm-language}, Kalchbrenner \etal \cite{seq2seq} led the way for the development of Encoder-Decoder architectures for mapping one sequence to another. This also has brought several advancements in Neural Machine Translation. Encoder-Decoder architectures encode a sequence into a fixed-sized dimensional space. It may be called latent space and every latent representation is decoded into a sequence. In addition to that, Sutskever el al. \cite{seq2seq2rnn} proposed to use separate RNNs for encoders and decoders. As a result, encoders are specialized in source languages and decoders learn to construct sentences in target languages. This framework has become the state-of-the-art method for translation tasks. Later, multisource NMT is proposed by \cite{multisource-nmt}. A separate encoder is assigned to learn each source language and it is shown that Encoder-Decoder networks can be extended by improving the state-of-the-art results.

\par DL approaches enable us to train models large models with automatic differentiation. It is an open question of what the machines really learn. These models sometimes cannot focus on important parts of inputs. Mnih \etal \cite{rnn-attention} imposes an attention model in RNNs inspired by human eyes. This method tries to sum numbers shown in images by focusing on the different parts of them at each step. It is shown that the model learns to eliminate unnecessary parts of images thanks to the additional hard attention mechanism. Later, this approach is adopted by image captioning research. Xu \etal \cite{sat} compares hard attention mechanisms with soft attention ones by revealing where the models focus on in describing an image. Soft attention creates a context from different parts while hard attention uses only one part at a time. Later, attention-based models are also adapted into video captioning and successful results are obtained \cite{video-captioning}.

\par Later, those attention-based models attract NMT researchers as a fixed-sized latent dimension is not enough to encapsulate all of the information in sentences. This leads to the development of Attentional RNN Encoder-Decoder architectures. Bahdanau \etal proposes an attentional sequence-to-sequence model that generates a different context vector from a sentence at each time. Simply, the architecture is designed to focus on different input words while generating each target word. This is similar to the human translation approach. After extracting the context of the sentence, we can decode words considering relevant parts in the source sentence. In DL perspective, it provides relaxation to the network that improves learning by refining irrelevant parts in sentences. Later, the work in \cite{luong-etal-2015-effective} improves Bahdanau Attention mechanism and proposes a local attention method in addition to the global one. The network takes all encoder outputs into account in the global attention while the local attention chooses a position and extracts the context vector from a window near the position. It is very similar to the hard attention proposed by \cite{rnn-attention}, but this method is fully differential. Therefore, it does not require further Bayesian optimization techniques.

\par Contrary to the previous approaches, Vaswani \etal \cite{transformer} proposes Transformer constructed with self-attention units instead of RNN units. They preserve Encoder-Decoder architectures and self-attention units reveal the relation between input words to understand the word's real meaning in the given sentence. This approach uses feed-forward networks therefore an extra positional indicator is needed. Otherwise, the network has no notion of the order in the sequence. We know that the order of words plays a crucial role in language modeling. Therefore, they feed the position of the words with sine and cosine functions. This architecture has become state-of-the-art in NMT as well as many NLP applications \cite{bert}.

\par Based on those advancements, Neural Sign Language Translation research has emerged. Camgoz \etal \cite{camgoz2018neural} formalizes the first architecture with the attention-based models. The architecture requires an additional effort for obtaining tokens. In NMT, tokens are produced with words or characters, but it is not clear to obtain tokens for sign language translation. The work introduced three different tokenization variants. The first one is gloss to sentence translation which is very similar to spoken language translation. The Second is sign to gloss to sentence translation. Sign to gloss is performed by another network. The network is needed to be trained separately on sign to gloss alignment. Therefore, they use the network proposed by Koller \etal \cite{re-sign}, but it is trained on the same dataset with the translation dataset. The last one is sign to sentence translation. Contrary to the first two variants, this converts frames to sentences without any explicit intermediate representation like glosses. It is formalized as two parts sequence-to-sequence model and tokenization part. The sequence-to-sequence model is adapted from NMT. Tokenization is the tricky part and requires further research. In the paper, a pre-trained CNN and NMT model is trained in tandem. the CNN extracts features from frames and features are concatenated in time. Naturally, the resulting model has longer sequences compared to the gloss-to-sentence model. But, the attention mechanism seems to handle the long sequences. 

\par In addition to that, Ko \etal \cite{korean} proposes a new tokenization method that uses human keypoint coordinates instead of CNN last layer features. Later, Sign Language Transformers are proposed by \cite{Camgz2020SignLT}. In this study, glosses and sentences are learned jointly, yet requires tedious gloss annotations. It uses common representations of images as tokens. The result exceeds golden gloss to sentence translation. It is inferred that glosses are not perfect tokens for SLT. More relaxed representations can attain higher scores instead of hard representations like glosses.

\section{Neural Sign Language Translation Datasets}
\label{list:nslt-dataset}
\begin{figure}[htbp]
\begin{center}
\includegraphics[width=1\columnwidth]{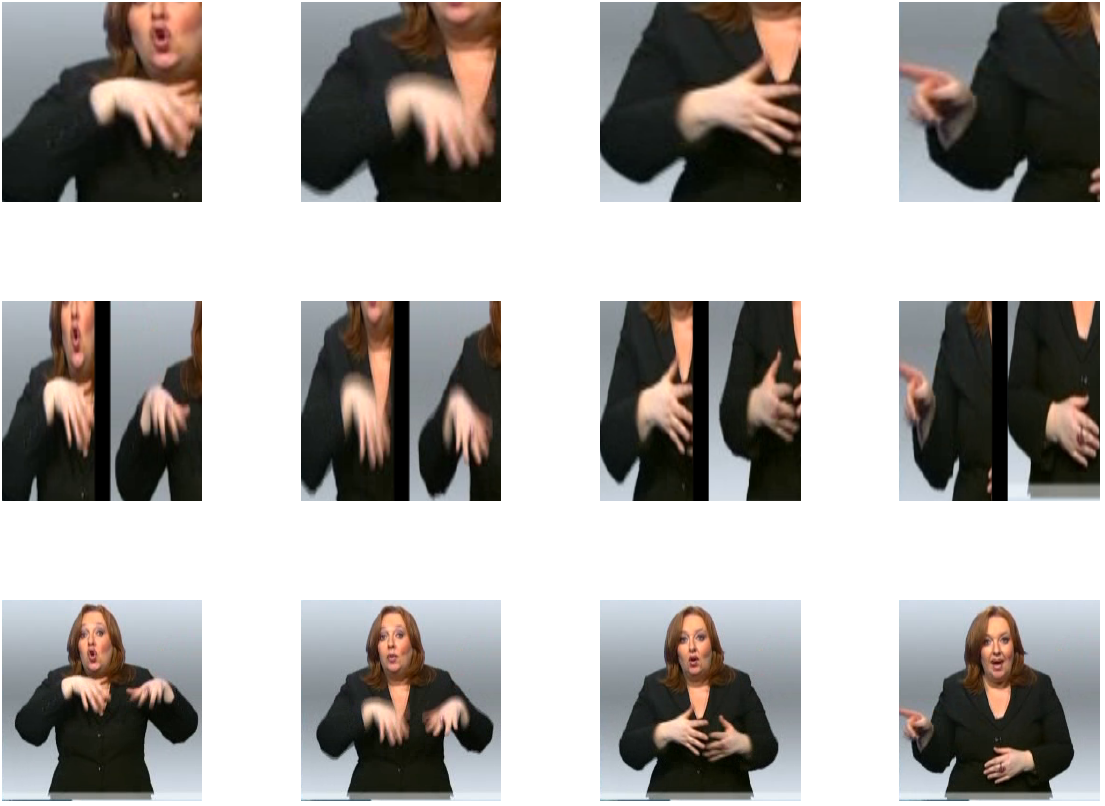}
\end{center}
\caption{Three sample images from right hands, both hands and full frame from top to bottom.}
\vskip\baselineskip
\label{fig:translate-imgs}
\end{figure}

	We use RWTHPHOENIXWeather 2014T introduced by Camgoz \etal \cite{camgoz2018neural}. It is an extended version of RWTHPHOENIX-Weather 2014. This is the most challenging SLT dataset publicly available to the best of our knowledge. They published this dataset with full-frames that involve faces, hands and upper-body parts of signers. We add two other settings. We identify body parts with OpenPose \cite{cao2018openpose},\cite{cao2017realtime}, \cite{simon2017hand}. For the first settings, we use only right hand crops, concatenation of left and right hands are prepared for the other. The samples is seen in Figure \ref{fig:translate-imgs}.
	 
The dataset is divided into train, test and validation. The train consists of 7096 video and sentence pairs. Validation and test data sets consist of 519 and 642 samples, respectively. The unique word size is 2891 and the unique gloss size is 1235.

\chapter{LEARNING TOKENIZATION}
\label{chapter:learning-token}
In this section, we present the overall architecture of Neural Sign Language Translation. We propose different tokenization methods to obtain string sign embeddings. First, we use a 2D-CNN trained for a task relevant to SLR. As a prior knowledge, we know that hand shapes are very important to understand what signers say. The literature suggests that possible hand shapes are similar in all sign languages, but their meanings differ from language to language. Therefore, learning hand shapes may be beneficial to obtain a generic tokenization layer. Secondly, 3D-CNNs have become very successful in action recognition tasks. We adapt a 3D-CNN into the tokenization layer to enable knowledge transfer from action recognition. The major advantage is that sign language videos involve so many redundant frames that temporal convolution may decrease training time by eliminating unnecessary information.
\section{Neural Sign Language Translation}
\label{list:nslt}
\begin{figure}[htbp]
\begin{center}
\includegraphics[width=1\columnwidth]{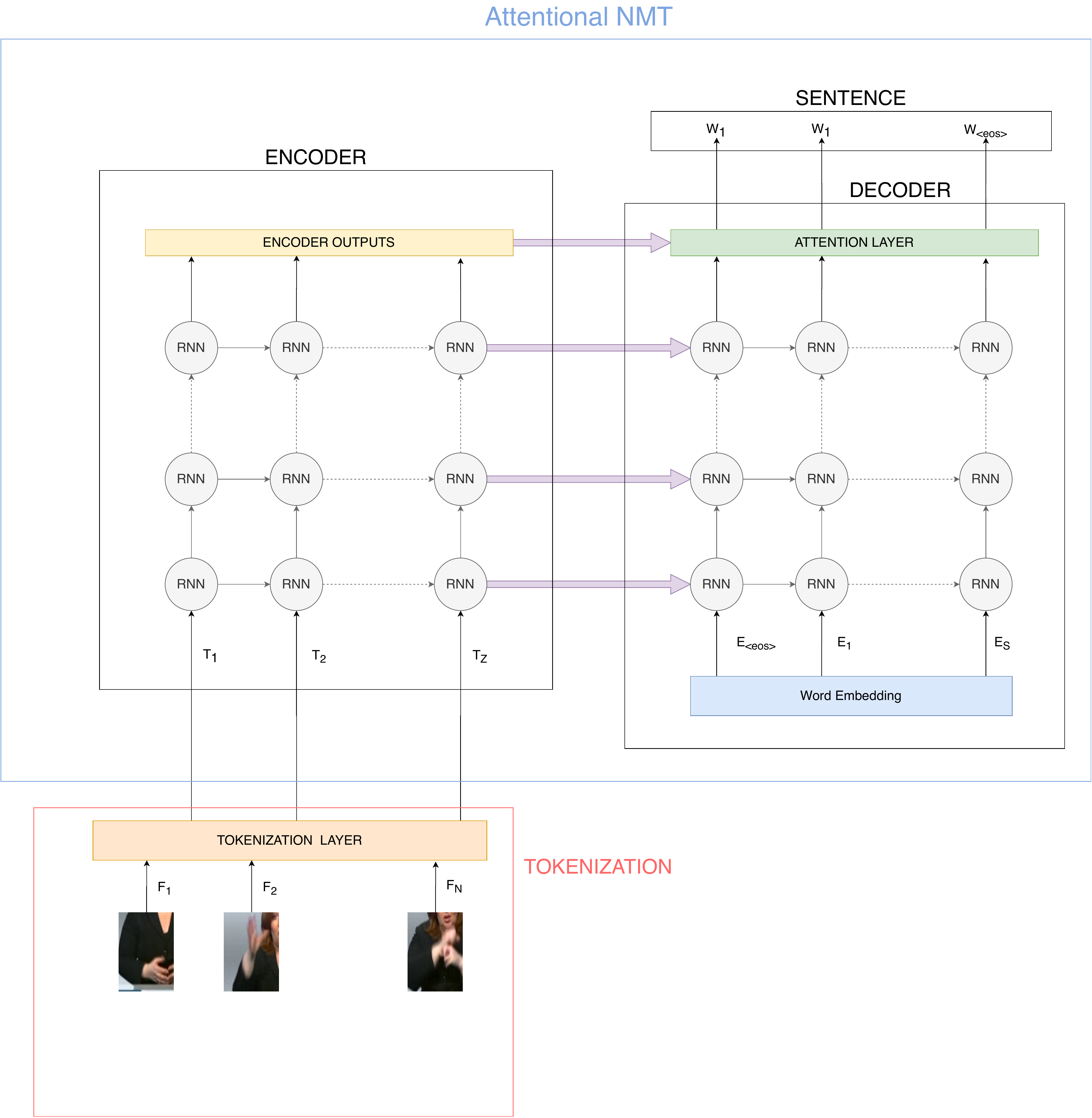}
\end{center}
\caption{Overall schema of Neural Sign Language Translation.}
\vskip\baselineskip
\label{fig:nslt-schema}
\end{figure}

	In this section, we introduce overall NSLT schema which mainly consists of two parts, tokenization and Neural Machine Translation (NMT). NMT handles sequence to sequence modelling by mapping input tokens to output words. When the sequence length is very long, attention mechanisms tackle long term dependencies. There are three major choices  for attention methods, Bahdanau \cite{bahdanau}, Luong \cite{luong-etal-2015-effective} Attention and Self-Attention \cite{transformer}. The first two are designed to function on LSTMs, GRUs \cite{gru} or Vanilla RNNs and the other requires a novel architecture called Transformers. The parts of NLST is seen in Figure \ref{fig:nslt-schema}.

\subsection{Sign Embeddings}
\par We think the tokenization part is essential to adopt new NMT techniques into SLT. If we find appropriate representations of videos, the developed methods in NMT are effectively applied. Different from spoken language translation, the tokenization part in NSLT also deals with the visual properties of SLs. Therefore, the tokenization part does not have to produce discrete tokens. In NMT, sentences are represented with discrete words or n-grams. Those tokens are converted to continuous space to provide more semantic information. However, in NSLT, the raw input is already continuous. The frame-to-gloss tokenization approach extracts discrete tokens, glosses, from videos to mimic successful NMT systems. We claim that this process can cause information loss. A transformation from continuous to discrete space is not required as they are again transformed into continuous space eventually. Also, small perturbations in discrete domains may amplify errors in sentence levels.

\par Our approach is based on strong sign embeddings. A sign can be represented with a single vector or concatenation of several frame representations. Additionally, the last layer of 3D-CNNs is also a good candidate as signs consist of consecutive frames. It is important that our tokenization approach does not require to represent strictly signs as the NMT module can combine distributed information implicitly. Therefore, we extend the scope of tokenization and our frame-level tokenization produces sign level embeddings to be replaced with word embeddings in NMT. Conversely, sign embeddings can be learned from other data sources. The word embedding is a single dense layer, but tokenization to sign embeddings should involve CNN like architectures. As they are very deep, they cannot be learned easily with sentence-video pairs. We know that SL research generally suffers from limited data. Considering all the issues, we claim that frame-level tokenization is a more proper approach and enables additional supervision, unlike gloss level tokenization. In the following subsections, we propose different frame-level tokenization approaches and compare them.
\subsection{Attentional NMT}
\label{list:diff}
Assume that we have an optimal tokenization mechanism that produces $Z$ tokens, $t_{1:Z}$, for a given $N$ length frame sequence, $x_{1:N}$. Note that sequences can be reversed after the tokenization, $\psi$, as suggested in \cite{seq2seq2rnn} to shorten the average distance between input tokens and decoded words. This helps to decrease long term dependencies while decoding. The problem is reduced to learning a sequence to sequence model between $t_{1:Z} \rightarrow w_{1:S}$ where $w_{1:S}$ is the sequence of target words. Encoder-Decoder, $g \circ f$, architecture is used to model this mapping between input and target. The encoder, $f$,  produces outputs, $o_{1:Z}$ and hidden states, $h_f$. The outputs are used to generate more meaningful contexts. The hidden states are the only way to bridge the encoder and decoder in the absence of an attention mechanism. The encoder produces the same number of outputs as the number of tokens. Also, the last hidden states are used to initialize the hidden states of the decoder. In this manner, the context can be conveyed to the decoder. In short, the encoder produces $Z$ outputs and the same number of hidden states as the encoder layer size. 

\par The resulting $o_{1:Z}$ is used in attention mechanisms. The decoder, $g$, generates its own outputs, $h_{1:S}$ and produces the target words, $w_{1:S}$, by using attention mechanisms. The attention mechanism develops a similarity notion between the encoder outputs, $o_{1:Z}$  and the decoder outputs, $h_{1:S}$. The decoder is designed to provide the most likely target word sequence if a token sequence is given. To construct a proper sentence, the decoder uses the encoder outputs,
$h_{1:t}$, the generated words, ${w}_{1:t}$  and all the encoder outputs, $o_{1:Z}$ to generate the word at step $t+1$. Eventually, the network decomposes the conditional probability of a word at a time. In the following equations, the Encoder-Decoder formulation is stated:
\begin{align}
	t_{1:Z} &= \psi(x_{1:N}) , \\
	o_{1:Z},h_{f} &= f (t_{1:Z}) ,	\\
	w_{1:S}, h_{1:S} &= g(o_{1:Z},h_f) , \\
	p(w_i|w_{<i}, t_{1:Z} ) &= g \circ f (t_{1:Z}) , \\
    \log{p(w_{1:S}|t_{1:Z})} &= \sum_{i=1}^{Z}\log{p(w_i|w_{<i},o_{1:Z})} .
\end{align}
\par
\begin{figure}[htbp]
\begin{center}
\includegraphics[width=1\columnwidth]{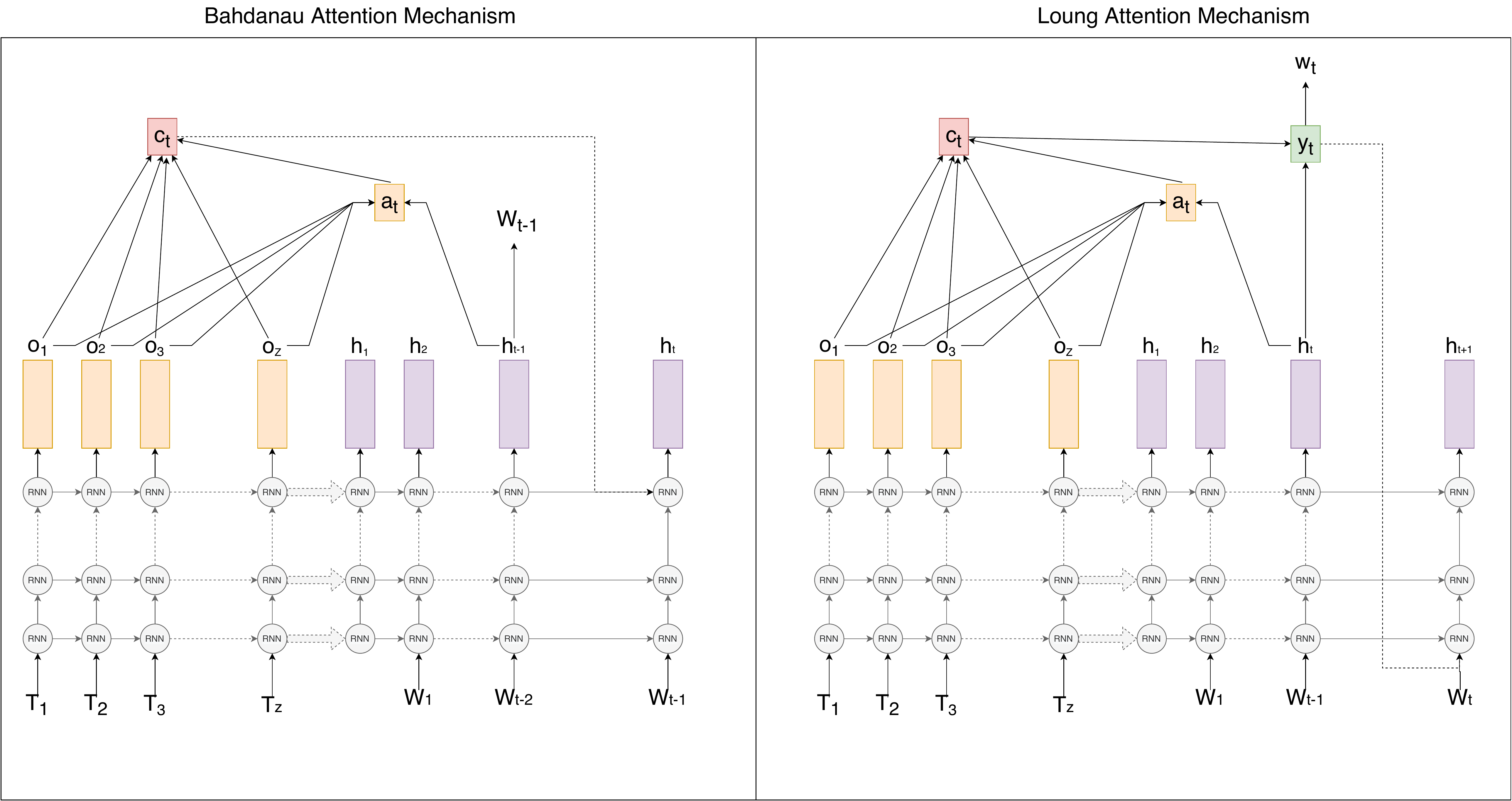}
\end{center}
\caption{Illustration of Bahdanau and Luong attention mechanisms.}
\vskip\baselineskip
\label{fig:attentions}
\end{figure}

 There are two main attention mechanisms when the Encoder-Decoder architecture is constructed with recurrent neural networks. The first one is Bahdanau and the second is Luong attention mechanism.
Figure \ref{fig:attentions} displays the overall architectures of the two attention mechanisms. Bahdanau attention creates context representations and generates the words as follows:
\begin{align}
	e_{1:z} &= \alpha(h_{t-1}, o_{1:z} ) ,\\
	a_{i} &= \frac{\exp(e_i)}{\sum^{z}_{k=1}\exp(e_k)} \CommaPunct \\
	c_t &= \sum_{i=1}^{z} a_i o_i ,
\end{align}
where $c_t$ is fed into the last layer of RNN cell. $\alpha$ is an arbitrary distance function between two vectors.

\par Luong Attention mechanism obeys the following equations. $\alpha$ is an arbitrary distance function between two vectors. $\theta$ is a dense layer with softmax. $y_t$ is fed into the first layer of the decoder with the word embedding such that
\begin{align}
	e_{1:z} &= \alpha(h_{t}, o_{1:z} ) , \\
	a_{i} &= \frac{\exp(e_i)}{\sum^{z}_{k=1}\exp(e_k)} \CommaPunct \\
	c_t &= \sum_{i=1}^{z} a_i o_i , \\
	y_t &= \theta(c_t, h_t)	.
\end{align}

\par The main difference between them is that Bahdanau uses the decoder output of the previous time step and the other uses the current decoder output. The second difference is that Luong feeds its decision over attention, $y_t$ into the first layer.

\section{Tokenization with 2D-CNN Trained on Hand Shape Classification}
\label{list:2D-CNN}

\begin{figure}[htbp]
\begin{center}
\includegraphics[width=1\columnwidth]{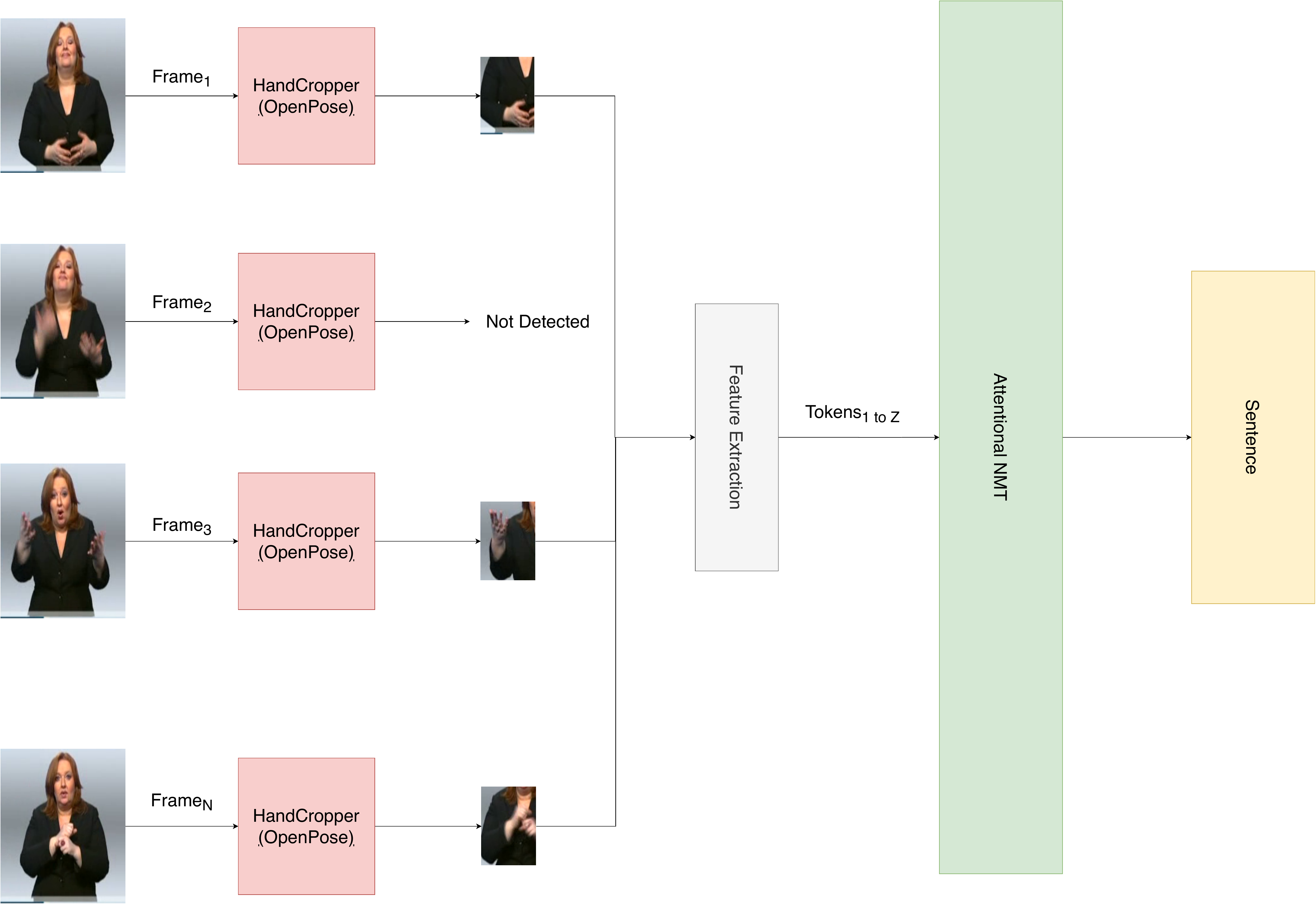}
\end{center}
\caption{Tokenization with a 2D-CNN trained for hand shape classification.}
\vskip\baselineskip
\label{fig:tokenization-handshape}
\end{figure}

	Hand shapes play a crucial role in interpreting the meaning of sign languages. We claim that a tokenization layer can benefit from this prior knowledge. Firstly, hand crops should be extracted to be processed for spatio-embeddings. There are several options for this challenge. A hand tracker or a hand detection network may be deployed. Openpose \cite{simon2017hand}, \cite{cao2017realtime}, \cite{cao2018openpose} is a strong human keypoint detector functioning in different conditions. Also, Ko \etal \cite{korean} shows it is very successful as a tokenization layer with a different formulation. OpenPose naturally eliminates some frames as it is not able to detect hands in blurry images. This effect may decrease the noise level in training but can result in ignoring some important frames. We consider this very unlikely as the training set of OpenPose includes high variation in terms of light conditions and camera views. The overall architecture is seen in Figure \ref{fig:tokenization-handshape}.
	
\par Koller \etal \cite{deephand} introduces a big dataset and a method dealing with weak labeling. Most of the dataset is common with the translation dataset. We suspect that a network trained on this method and dataset may not be robust for datasets from other domains. Therefore, we propose two other solutions to deal with that while maintaining domain independence of the network. The first one is training a network on multiple tasks and datasets. It may force the tokenization layer to learn a more general feature space. The second is using domain adaptation techniques explicitly. If there is some data from the translation dataset, this technique may enable the feature extractor network to learn domain-independent features. In this manner, we may have an improved tokenization layer for unknown sign language data.

\subsection{Multitask Learning}
\label{list:multitask-learning}
\begin{figure}[htbp]
\begin{center}
\includegraphics[width=1\columnwidth]{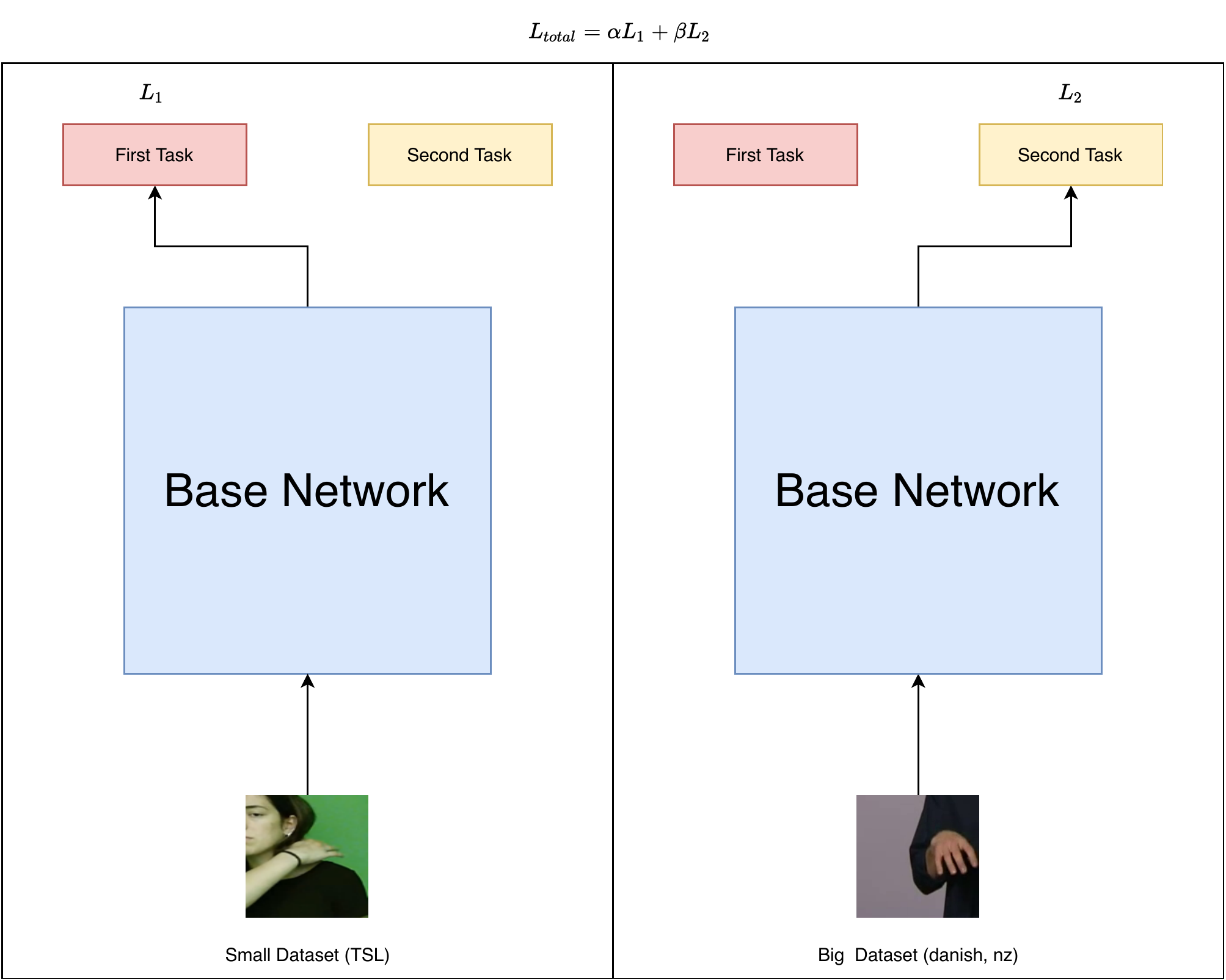}
\end{center}
\caption{Multitask learning framework.}
\vskip\baselineskip
\label{fig:multitask}
\end{figure}

	The aim of using Multitask Learning is to deal with two main challenges in SLR. The first one is that data is weakly labeled. It leads the feature extractor network to learn a bad latent representation space. The second is that the environmental conditions where data is collected make the network very sensitive to perturbations in input space. Small changes in the background or camera angles can lead the network to be saturated. The network cannot produce robust features due to this. As seen in Figure \ref{fig:multitask}, the architecture is designed to obtain a common latent representation for different datasets.
	
\par As mentioned in Section \ref{list:slr-dataset}, we have different data sources and different labeling approaches. Furthermore, the datasets vary in terms of the noise level. They may not be seen as sufficiently strong signals to train a good tokenization layer if used alone. However, we claim that multitask learning is a proper way to combine those weak signals to train a strong network. Therefore, the TSL dataset may be used as a regularization factor on training danish and NZ datasets since the TSL dataset is very small. This may provide two different advantages. Firstly, it may reduce the negative effect of incorrect labels in the big dataset as the TSL dataset is more reliable in terms of annotation. Therefore, the TSL dataset imposes a regularization effect on the network which prevents it from overfitting to the noisy labels. The second advantage is that its labels involve contextual information meanwhile labeling the big dataset only concern about context-independent hand shapes. In other words,  the small data involves syntactic and semantic meaning contrary to the big. 
\par Another major benefit is that combining different datasets naturally makes the network robust to different environment conditions since they increase the diversity of signers, background and camera settings. We expect that this approach attains better results without any information from the target domain.
\par However, multitask learning requires additional effort to attain a balanced learning schedule between tasks. Assume that we have two different tasks and we update weights after one batch from two tasks are iterated. The loss function is
\begin{equation}
	L_{\textit{total}} = L_1 + L_2 .
\end{equation}
This means that both tasks are equally weighted. Generally, two tasks are unlikely to be learned at the same rate from a DL perspective. Moreover, sometimes loss function scales of separate tasks differ dramatically. Even if the loss function is the same, the number of classes can be different and that may generate a huge imbalance in the loss function. Therefore, the loss function may be arranged as
\begin{equation}
		L_{\textit{total}} = L_1 + \beta(t) L_2 .
\end{equation}
The coefficient of the second function may be dependent on the iteration number $t$ or a constant value. The first loss term's coefficient is always 1 as it can be arranged with the learning rate.
\subsection{Domain Adaptation}
\label{list:domain-adaptation}
\begin{figure}[htbp]
\begin{center}
\includegraphics[width=1\columnwidth]{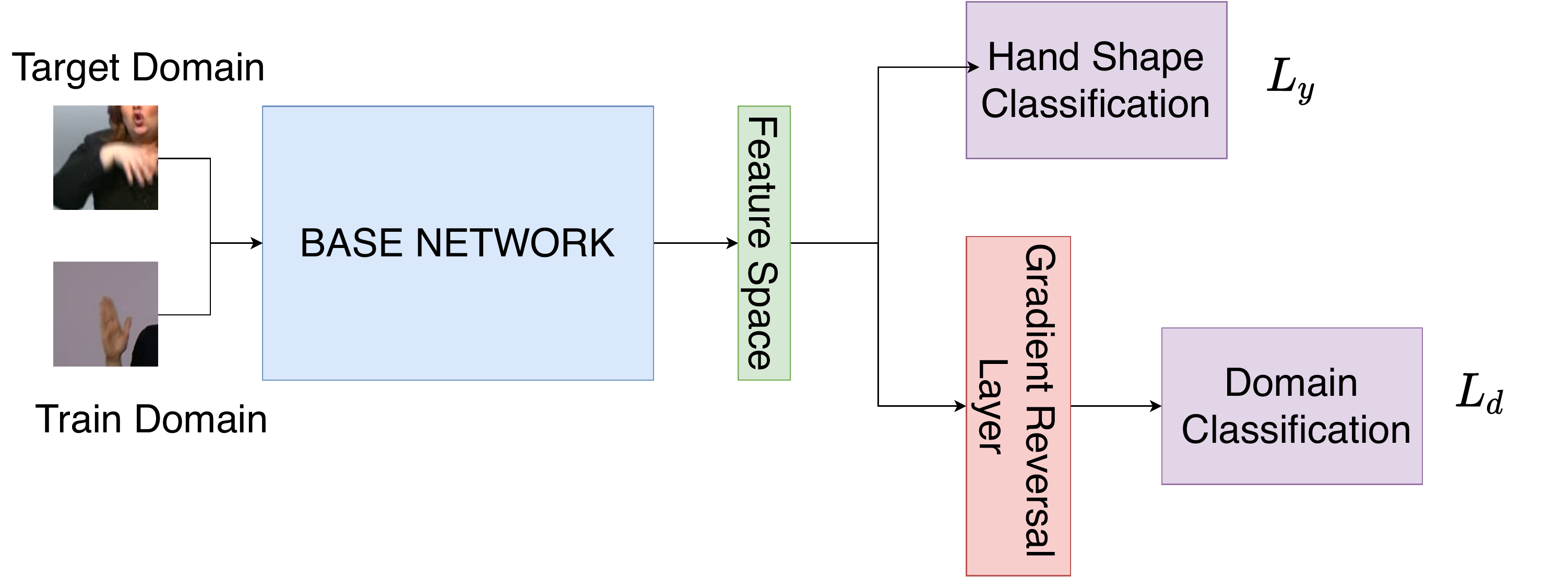}
\end{center}
\caption{Illustration of domain adaptation with Gradient Reversal Layer.}
\vskip\baselineskip
\label{fig:domain-adaptation}
\end{figure}

	We know that every sign language dataset involves nearly the same hand shapes, but some domain shifts occur between datasets. It can be fixed with domain adaptation techniques, however, some annotation or incorporation of synthetic data is needed. Ganin \etal \cite{domainadaptation} proposed an unsupervised method that can be used on deep architectures. It may be useful to apply this method while training the feature extractor network. This enables the network to learn domain invariant features while gaining the capability to distinguish the classes. We expect this approach to decrease classification error in the target domain. This may contribute to an increase in translation quality. The overall architecture is illustrated in Figure \ref{fig:domain-adaptation}.
	
	The gradient reversal layer (GRL) and domain classification layer are the modifications to the normal architecture. Labels of translation dataset frames are unknown, but their domain label is different than training data. Hence, the method may approximate the feature space distribution of the training data to the translation data. The gradient reversal layer is identity function in the forward pass. But, in the backward pass, it reverses the gradient with an amplifying factor such that
\begin{align}
	GRL(x) &= x , \\
	\frac{\delta GRL(x)}{\delta x } &= -\lambda(t) \cdot
\end{align}

\par This arrangement means that the learned features to distinguish the domain are diminished with a gradient reversal trick. The $\lambda$ is gradually increased from 0 to 1 as learning at the initial steps may be harmed by noise caused by the gradient reversal layer.

\section{Tokenization with 3D-CNN Trained on Action Recognition}
\label{list:3D-CNN}
Instead of spatio-embeddings with 2D-CNNs, 3D-CNNs can produce spatio-temporal embeddings as a tokenization layer. Carreira \etal \cite{I3D} proposes a novel 3D-CNN architecture providing promising results on human action recognition. Also,  the studies in \cite{3d1}, \cite{3d2} show that 3D-CNNs are suitable for videos in sign languages. 
\par
\begin{figure}[htbp]
\begin{center}
\includegraphics[width=1\columnwidth]{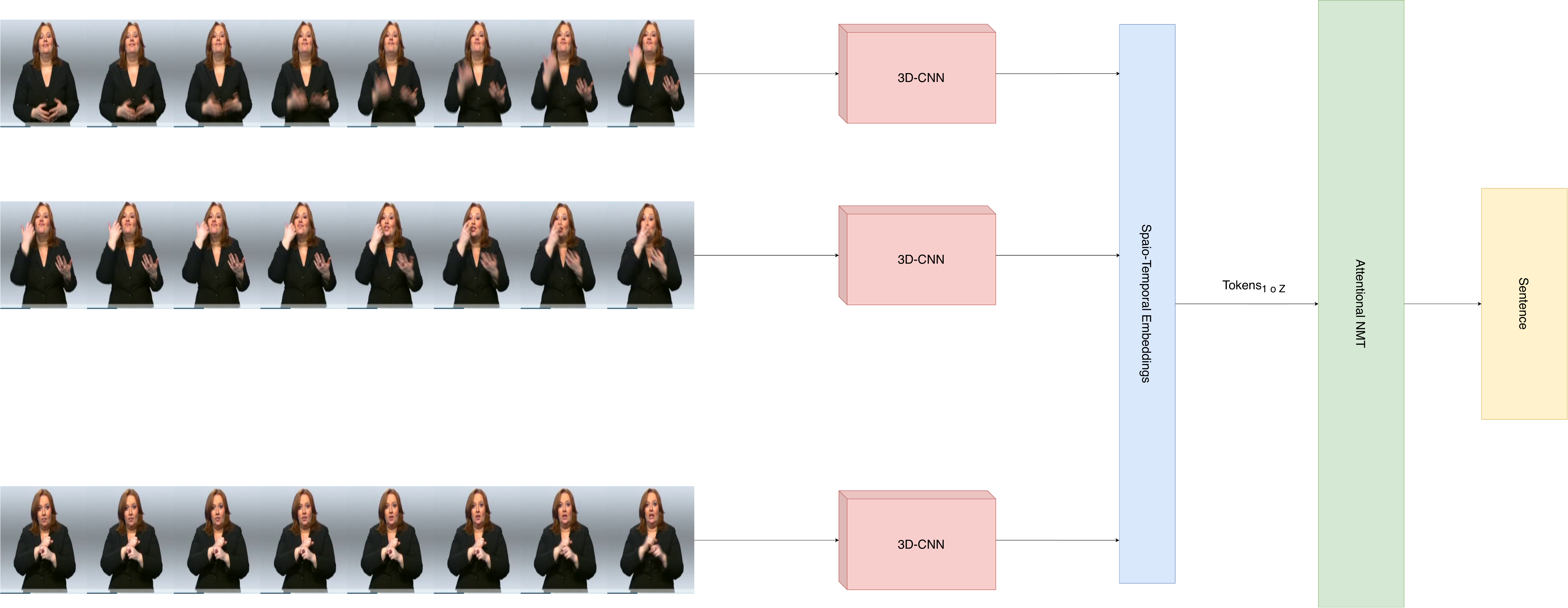}
\end{center}
\caption{Illustration of Tokenization with 3D-CNN.}
\vskip\baselineskip
\label{fig:3d-cnn}
\end{figure}

 We know that consecutive frames are very similar to each other. Therefore, temporal convolution may prevent sequence-to-sequence models from handling very long sequences. Furthermore, 3D-CNNs may enable information transfer from human action recognition. In this thesis, we emphasize that hand shapes are very important for sign languages. This method is able to take human action into account along with hand shapes. The overall mechanism can be seen in Figure \ref{fig:3d-cnn}.

\chapter{EXPERIMENTS \& RESULTS}
\label{list:learning-token}

For the translation dataset, we use the dataset introduced in Section \ref{list:nslt-dataset}. Tensorflow \cite{tensorflow} version of OpenNMT \cite{opennmt} library is used for sequence-to-sequence modeling. PyTorch \cite{pytorch} is utilized to import pre-trained models and construct customized networks. Also, end-to-end learning experiments are performed with the code published by Camgoz \etal \cite{camgoz2018neural}.  To evaluate our translation results, we used the same metrics in \cite{camgoz2018neural}, BLEU (BLEU-1, BLEU-2, BLEU-3, BLEU-4) \cite{bleu} and ROUGE (ROUGE-L F1-Score) \cite{rouge}, to enable consistent comparisons. 
\par Each model is trained up to 30.000 iterations. Models parameters are saved every 1000 iterations and resulting parameters are average of the last five saved parameters. This reduces biases in selecting the best models. Model hyperparameters are chosen to make the models as simple as possible to make the experiments generic. Concerning the computational effort to train a translation model, it is not possible to perform an extensive grid search on hyper-parameters. The hyper-parameters and other details are listed in the Appendix.
	
\section{Input Analysis of Implicit Tokenization Layer}
\label{list:end-to-end}
	We claim that the end-to-end learning scheme proposed by Camgoz \etal \cite{camgoz2018neural} is not effective due to small data size. The unique word size is 2891 and the unique gloss size is 1235 whereas there are a total of 7096 sentence-video pairs. Despite the fact that RWTH-PHOENIX-Weather 2014 is seen to be sufficiently large for SLR, SLT requires more sentence-video pairs. As a prior knowledge, we know that hand parts of frames involve the most crucial information to understand signers. However, sign languages are very context-dependent languages so general frame information is also important. 
\par We prepare three different versions of the translation data as illustrated in Figure \ref{fig:translate-imgs}. The first one is full frames, the second one is only right hand crops and the last one is the concatenation of both hands. The reaction of implicit tokenization to the different input types provides clues about the effectiveness of this method. If the method attains the best results with full frames, it is claimed that the implicit tokenization layer is able to obtain strong feedback signals from the sequence-to-sequence model. It means that the tokenization layer can infer the information both from hands and the context of the frame. 
\par AlexNet is used to extract spatio-embeddings. To test the claims above, we carry out six experiments. In the first three, the experiments are conducted with the end-to-end learning scheme, which means that AlexNet is fed with gradients coming from the sequence-to-sequence model. In the others, AlexNet is not tuned further with the NMT model.
\begin{table}[thbp]
\caption{Test results of implicit Tokenization.}
\label{table:endtoend}

\begin{center}
\begin{tabular}{|l||c|c|c|c|c|}
 \hline
   & ROUGE & BLEU-1 &  BLEU-2&  BLEU-3&  BLEU-4 \\
 \hline
 Full-Frame \& End-to-End  &   30.70  & 29.86   &  17.52  &  11.96   &  9.00  \\
 \hline
 Both Hands \& End-to-End & \textbf{34.53} &     \textbf{33.65}   &    \textbf{21.01} &     \textbf{15.19} &     \textbf{11.66}  \\
 \hline
 
 Right Hand \& End-to-End & 31.89 & 30.57  &  18.67 &  13.19 & 10.25   \\
 \hline
 \hline
  Full-Frame \& Only NMT  &   29.25  & 30.91   &  17.42  &  11.80   &  8.85  \\
 \hline
 Both-Hands \& Only NMT  &   30.17  & 30.53   &  17.83  &  12.32   &  9.53  \\
 \hline
 Right-Hand \& Only NMT  &   \textbf{31.77}  & \textbf{32.67} &  \textbf{19.76} &  \textbf{13.79}   &  \textbf{10.50}  \\
 \hline
\end{tabular}
\end{center}
\end{table}	
\par The results raise doubts about the effectiveness of implicit tokenization. Full frames give the worst result while using both hand crops outperform the others drastically as seen in Figure \ref{table:endtoend}. It is interesting that tuning the tokenization layer has nearly no effect on full frames and right hands. Despite that, the tokenization layer is improved with both hands where BLEU-4 score increases from 9.53 to 11.66. The reason is that both hands have an unnatural discontinuity which is very distinct from images in ImageNet. This effect prevents the tokenization from being stuck in a local optimum.
\par This result suggests that feedback signals coming from the sequence-to-sequence model are not sufficient to train a strong tokenization layer. However, the tokenization layer is suitable to be improved with external information. Moreover, the end-to-end learning needs a huge amount of computational power while sustaining no positive effect except both hands setting. The end-to-end training requires 50 times more resources to converge within the same duration as the frozen one. In the right hand setting, it is also inferred that the tokenization layer cannot be specialized on hand shapes with the joint learning approach.
\section{Tokenization with a Frozen CNN}
\label{list:token-2d}

\begin{figure}[htbp]
\begin{center}
\includegraphics[width=1\columnwidth]{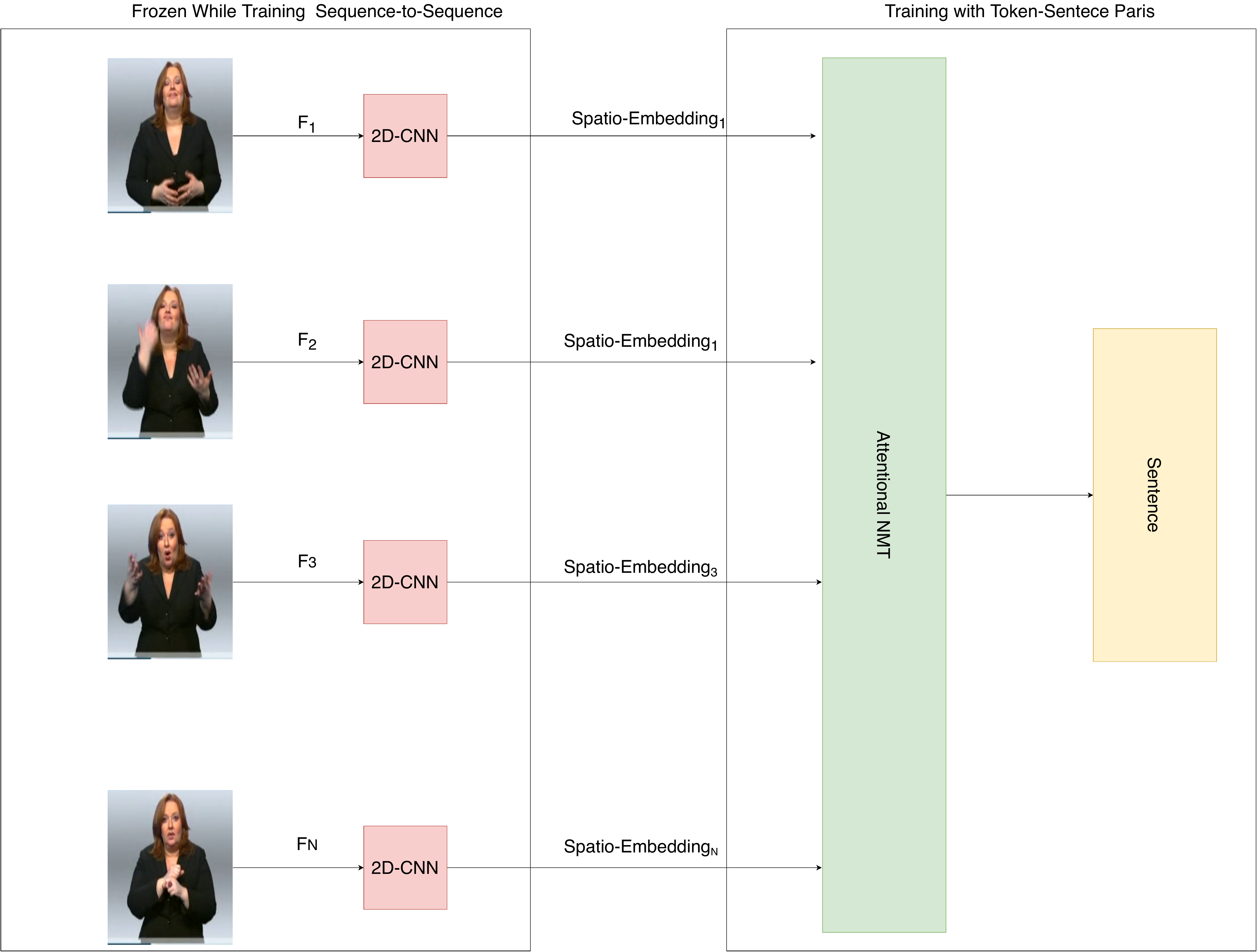}
\end{center}
\caption{Illustration of Tokenization with frozen 2D-CNN.}
\vskip\baselineskip
\label{fig:frozen-2d}
\end{figure}

	In this section, we aim to compare different frame-level tokenization approaches. First, we find the most successful attention mechanism to conduct the rest of the experiments. Later, we begin with a frozen 2D-CNN trained on ImageNet. We choose a more recent and efficient architecture instead of AlexNet. These experiments are considered as baseline. We expect that the success of the tokenization layer increases as we utilize various prior knowledge about sign languages. We test human keypoint features on our dataset with different human body parts. Furthermore, we check the validity of our approaches. Lastly, we test if human action recognition is related to sign language translation by using a 3D-CNN as a tokenization layer. Illustration of the overall schema is seen in Figure \ref{fig:frozen-2d}.

\subsection{Attention Search}
\label{list:attention-search}
	We conduct experiments to find the most appropriate attention mechanism. There are three options, Bahdanau, Luong and Self-Attention.  We choose a 2D-CNN trained on ImageNet as it is the most basic form for a tokenization layer. If a more sophisticated tokenization layer is used, pure contributions by attention mechanisms may not be observed. Inception v3 \cite{inception} is used without the classification layer as the base network. Additionally, full-frame input setting is used as it involves all human body parts compared to other settings, right hands and both hands. For each of them, the most accurate hyper-parameters are searched. For the first two, the search is initiated with the parameters suggested by \cite{camgoz2018neural}. Self-attention is applied for the first time in this dataset. Therefore, we start the search with the parameters in \cite{transformer}. The schema of this method can be seen in Figure \ref{fig:frozen-2d}. 
\begin{table}[thbp]
\caption{Comparison of  attention mechanisms.}
\label{table:attention-search}

\begin{center}
\begin{tabular}[t]{|l||c|c|c|c|c|}
 \hline
  Model & ROUGE & BLEU-1 &  BLEU-2&  BLEU-3&  BLEU-4 \\
 \hline
  Bahdanau & \textbf{29.41} & \textbf{30.46}  &   \textbf{17.79} &  \textbf{12.36} &  \textbf{9.40}  \\
 \hline
 Transformer  & 28.20 & 28.66  & 16.33 &  11.15 &  8.43  \\
 \hline
  Luong  & 26.94 & 27.46  &  15.12  &  10.38  &  7.93  \\
 \hline
 
\end{tabular}
\end{center}
\end{table}

\begin{figure}[htbp]
\begin{center}

\includegraphics[width=1\columnwidth]{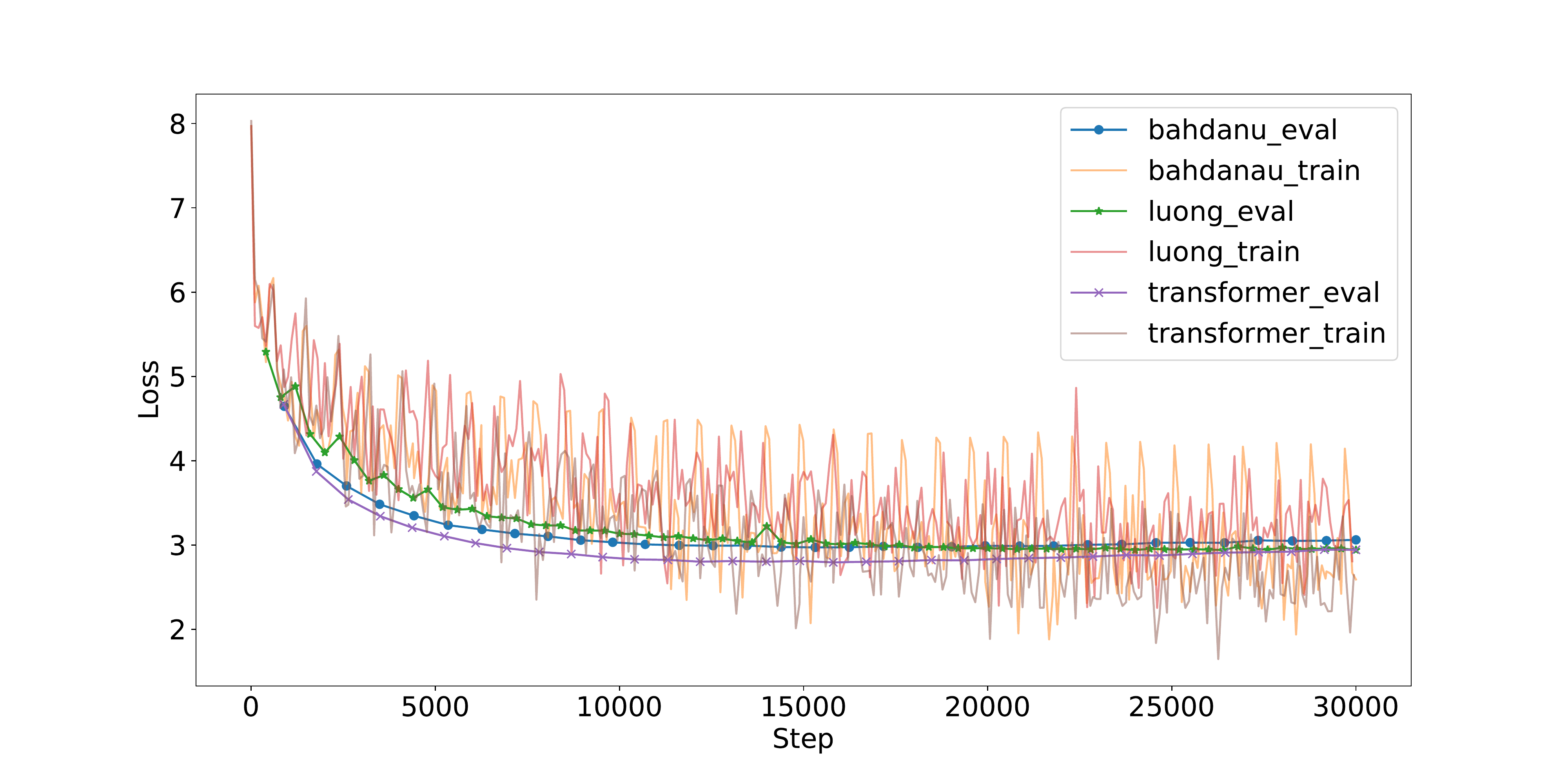}
\end{center}
\caption{Convergence graph of the models. }
\vskip\baselineskip
\label{fig:loss-graph}
\end{figure}

\begin{figure}[htbp]
\begin{center}
\includegraphics[width=1\columnwidth]{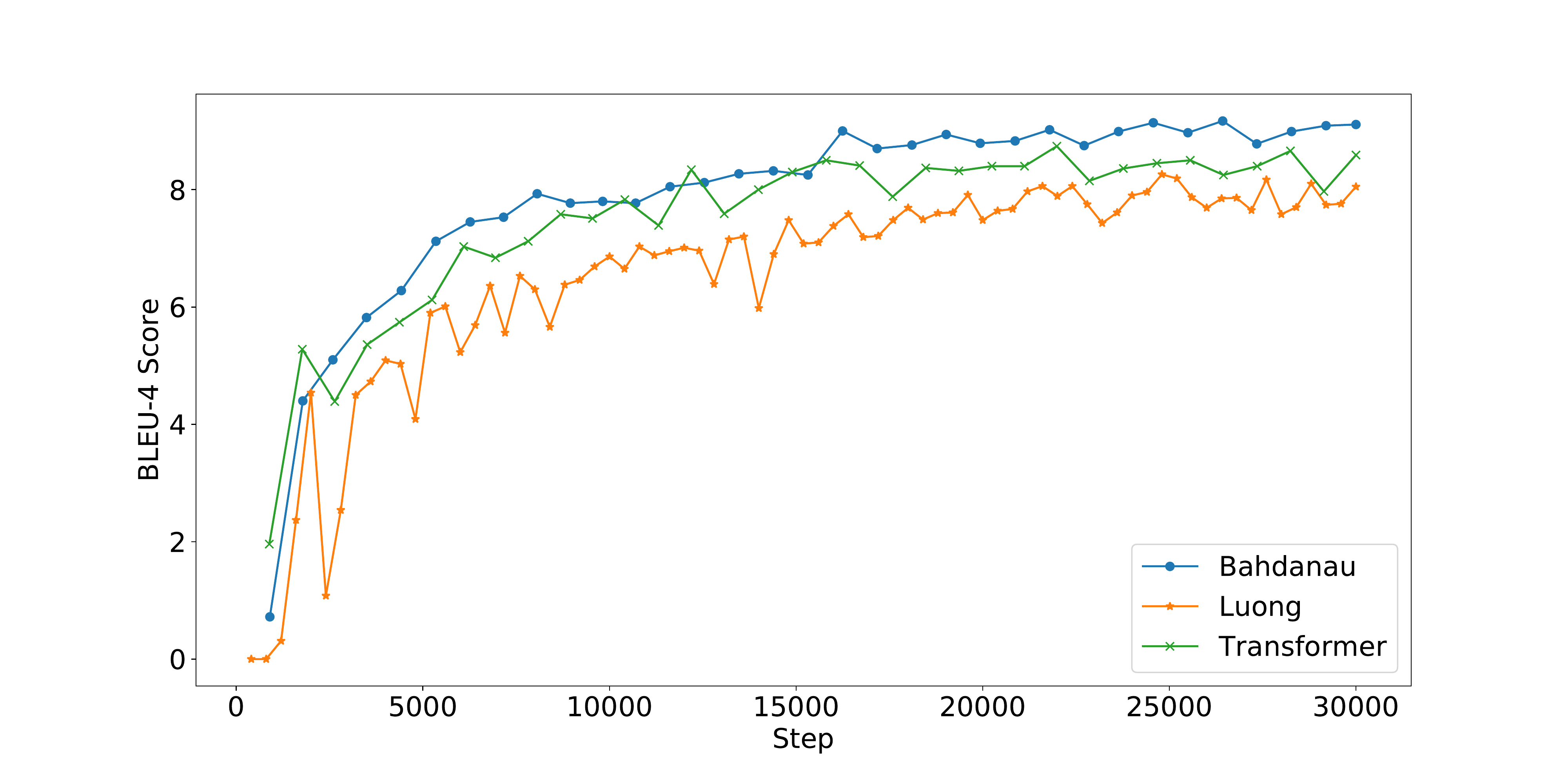}
\end{center}
\caption{BLEU-4 evaluation score graph of the models.}
\vskip\baselineskip
\label{fig:external-bleu}
\end{figure}

\par All of the models are converged and provide meaningful translations. Evaluation and training losses are depicted in Figure in \ref{fig:loss-graph}. The results indicate that Bahdanau outperforms others in all metrics as seen in Table \ref{table:attention-search}. The most important indicator is the BLEU-4 score. Evaluation set BLUE-4 scores are seen in Figure \ref{fig:external-bleu}. These results imply the same conclusion with test results. This result is consistent with our claim that large models can memorize the dataset easily. NMT with Bahdanau consists of fewer parameters compared to Luong and Self-Attention model.

\subsection{Transfer Learning From ImageNet}
\label{list:transfer-imagenet}
	Inception v3 is our base feature extractor network. Without further training and any modification, spatio-embeddings before the classification layer is used as the tokenization layer. These experiments are aimed to determine the baseline of tokenization for the rest of the experiments. Imagenet is a very diverse dataset consisting of 1000 classes and three million images. Most of the applications in Computer Vision networks are initialized with parameters inferred from the ImageNet dataset. The result may help us to understand which parts of frames are important since it is considered as the most general feature extractor.
	
\begin{table}[thbp]
\begin{center}
\caption{ Test results of different input settings.}
\label{table:imagenet}

\begin{tabular}[t]{|l||c|c|c|c|c|}
 \hline
    & ROUGE & BLEU-1 &  BLEU-2&  BLEU-3&  BLEU-4 \\
 \hline
 Full Frames & 29.41 &   \bftab 30.46  & \bftab   17.79 &  \bftab   12.36 &  \bftab  9.40   \\
 \hline
 
  Both Hands &  \bftab  29.62 & 30.24 & 17.46  &  12.14&  9.33  \\
 \hline
 
 Right Hands & 29.09 & 29.51  &  17.03 &  11.79 & 9.06   \\
 \hline

\end{tabular}
\end{center}
\end{table}

\par Contrary to the first three rows of Table \ref{table:endtoend}, the result in in Table \ref{table:imagenet} indicates that inputting full frames is the most effective approach. However, there are no big differences between the approaches as illustrated in Table \ref{table:imagenet}. 9.40 BLEU-4 score is attained without further effort compared to the result in \cite{camgoz2018neural} (9.58 BLEU-4)

\par The results are in contradiction with the last three rows of Table \ref{table:endtoend}: The right hands are the most successful with AlexNet whereas full frames attain the best scores with Inception v3. AlexNet has 4096 feature size and Inception has 2048. Inception architecture is designed with parallel filters that combine different information from varying windows sizes. AlexNet architecture has three fully connected layers that make overfitting very likely. Hence, Inception can utilize all information from full frames and exceeds AlexNet scores (8.85) in a full-frame setting.
\subsection{Human Keypoints}
\label{list:human-keypoints}
	OpenPose is a very robust keypoint extractor from both human bodies, faces and hands. Human keypoints may be converted into tokens. Ko \etal \cite{korean} suggests a pre-processing method instead of using keypoints directly. The method is as follows
	\begin{align}
		T_x = \frac{V_x - \hat{V_x}}{\sigma(V_x)} \\
		T_y = \frac{V_y - \hat{V_y}}{\sigma(V_y)} 
	\end{align}
where $V$ is the keypoint vector. The process describes the separate standard normalization of x and y coordinates. 
\par
\begin{figure}[htbp]
\begin{center}
\includegraphics[width=1\columnwidth,height=18cm]{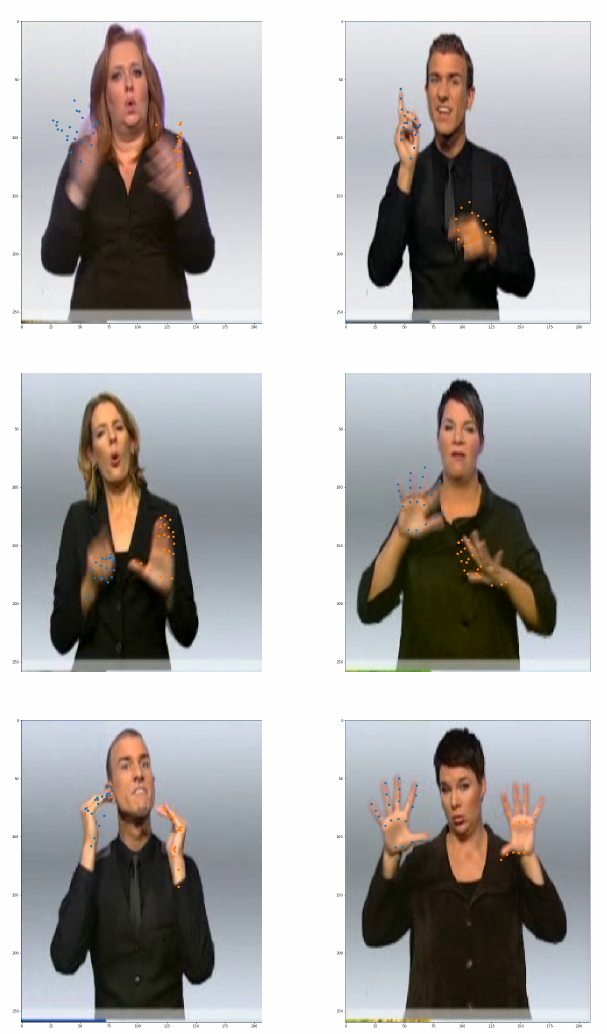}
\end{center}
\caption{Samples from OpenPose output coordinates.}
\vskip\baselineskip
\label{fig:openpose}
\end{figure}

 The upper body, right and left-hand keypoints are extracted from the translation dataset. To mimic the full frames setting, all of the keypoints are normalized within the body parts according to the method. This tokenization approach may have two advantages. The first one is that the training set of OpenPose is very challenging; so the method is applicable under a wide range of conditions. It automatically eliminates backgrounds and irrelevant parts of frames. Moreover, the interaction of the body parts is explicitly imposed. However, we know that subtle changes in hand shapes may cause big interpretation differences in sign languages. Therefore, this method may not be capable of identifying hand shapes since the outputs of finger positions are generally very noisy as seen in Figure \ref{fig:openpose}.

\begin{table}[thbp]
\label{table:keypoint}
\begin{center}
\caption{ Test results of Tokenization with keypoints.}
\begin{tabular}[t]{|l||c|c|c|c|c|}
 \hline
    & ROUGE & BLEU-1 &  BLEU-2&  BLEU-3&  BLEU-4 \\
       \hline

 Body \& Both Hands   & \bftab 32.85 & \bftab 33.18  & \bftab 20.39 &  \bftab 14.26 &  \bftab 10.92  \\
 \hline
 Both Hands  & 31.47 & 31.51  &  19.08 &  13.26  & 10.23  \\
 \hline
  Right Hands & 30.65  &  30.69  & 18.49 & 12.89 &  9.91  \\
 \hline
\end{tabular}
\end{center}
\end{table}

\par This method improves the baseline in every setting as seen in Figure \ref{table:keypoint}. Also, it is consistent with early findings that hands are the most important factor, yet other body parts are also helpful to understand signers. The third row shows that this method increases BLEU-4 by nearly 1 point compared to the last row of Table \ref{table:imagenet}. Therefore, keypoints are more accurate than the pre-trained Inception network to represent hand shapes. These findings emphasize that hand shapes should require special attention. A simple representation of hand shapes with coordinates gives promising results even if the OpenPose outputs involve considerable noise.
\subsection{Hand Shape Learning}
\label{list:hand-shape-learning}
	In this section, we want to show the effectiveness of hand shapes. The previous results show that any prior information about hand shapes improves the quality of translation. We adapt the proposed techniques in Chapter \ref{list:multitask-learning} to deal with weak annotation and domain shifts. We know that the big datasets are weakly labeled and small datasets do not cover most of the hand shape variations. Therefore, we train the network in a multitask manner. We check whether it enables the network to have a stronger feature space. First, we test this network on a small portion of the translation dataset. The test dataset is labeled manually by Koller \etal \cite{deephand}. Later, it is deployed as the feature extractor in Figure \ref{fig:tokenization-handshape}.
\par We also train a network with the unsupervised domain adaptation method explained in \ref{list:domain-adaptation}. The network is tested on the small test set to check if the technique is applied successfully. Later, the experiments reveal whether mitigating domains shifts has an important effect on translation.
\par
\begin{figure}[htbp]
\begin{center}
\includegraphics[width=1\columnwidth]{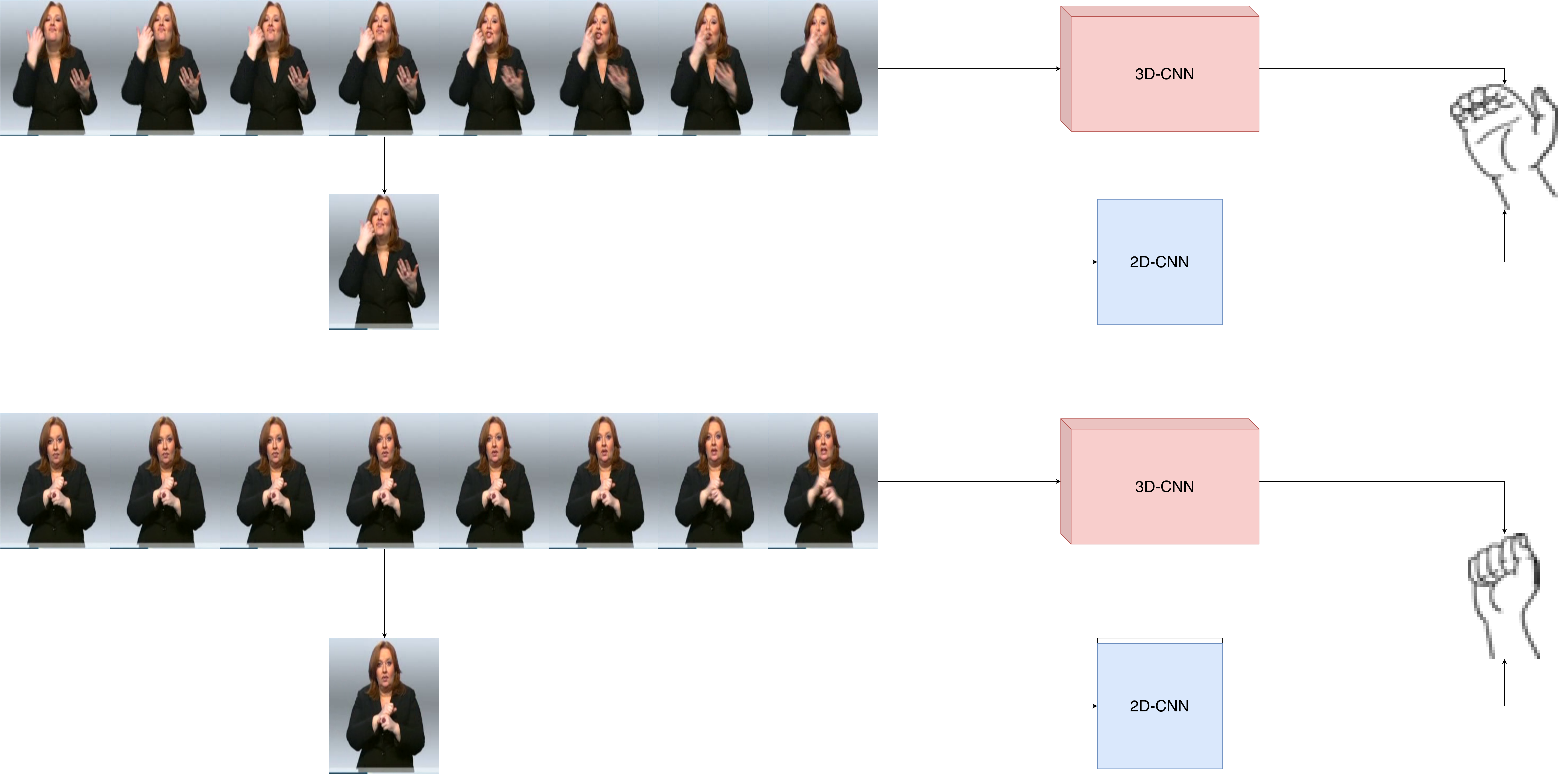}
\end{center}
\caption{Learning schema of hand shapes with a 3D-CNN.}
\vskip\baselineskip
\label{fig:3d-learning}
\end{figure}

 Lastly, we aim to find a way to reduce token sequences in length. Instead of sampling from frames, 3D-CNNs are applied to hand shape classification as seen in Figure \ref{fig:3d-learning}. As the data is coming from videos, we take a window around the desired hand shapes as inputs for 3D-CNNs. This technique may identify the desired hand shapes from sequences. We know that some frames are meaningful, but some are blurry and irrelevant in frame sequences. In these experiments, we use the big dataset with the architecture proposed by  
Tran \etal \cite{2d_1}. The network is initialized with pre-trained weights on Kinetics.

\par The datasets (New Zealand, Danish) in Section \ref{list:slr-dataset} are used to apply the techniques, expect German Sign Language, ph. The reason for the omission  is that the translation dataset is the modified version of ph. Therefore, it may harm our proposed approaches as training on those frames can cause memorization. We set up our experiment as if we do not have access to the target domain. We combine Danish and New Zealand datasets as the big dataset. The dataset from Turkish Sign Language is considered as the small dataset. Our baseline network is trained on only the big dataset to observe the effects of the proposed techniques.

\par The experiments are performed with only right hand crops in this section. We want to obverse the effectiveness of hand shape learning on tokenization layers. For simplicity, We ignore left hands for translation. The results reveal which methods provide additional supervision to the tokenization layer. Search for combining hand representations is left as future work. 

\par For consistency, the base CNN network is again Inception v3. For all tasks, the network is initiated with parameters trained on ImageNet to accelerate the training process. All the datasets are collected from sign language videos. Therefore, the distribution of hand shapes is highly imbalanced. We deploy a mini-batch sampler to overcome this challenge.

\begin{table}[b]
\begin{center}
\caption{Hand shape classification results on the the target domain.}
\label{table:cnn}

\begin{tabular}{|l||c|c|}
 \hline
   &  Top-1 &  Top-5   \\
  \hline
  Baseline  &  65.49  &  89.49    \\
 \hline
  Data Augmentation  &  67.07  &  90.65   \\
 \hline
  Multitask &  76.77  &  92.88   \\
 \hline
  Domain Adaptation  &  \bftab 78.74  &  \bftab 94.73   \\
 \hline
\end{tabular}
\end{center}
\end{table}

\par  Data augmentation in Table \ref{table:cnn} refers to the baseline method with random color jitter. The results indicate that Multitask Learning and Unsupervised Domain Adaptation increase top-1 accuracy to 11.28 and 13.25 points. It is indicated that there is a domain shift between the training domain and the translation domain. Also, Multitask learning enables knowledge transfer between the big and small datasets. It is clear that those approaches are adapted successfully.

\begin{table}[thbp]
\begin{center}
\caption{ Test results of Tokenization with hand shape representation.}
\label{table:hand-shape-tokenization}

\begin{tabular}[t]{|l||c|c|c|c|c|}
  \hline
      & ROUGE & BLEU-1 &  BLEU-2&  BLEU-3&  BLEU-4 \\
       \hline

  Big Dataset 2D-CNN & 34.59 & 35.52  &  22.37 &  15.80 &  12.17  \\
 \hline
   Big Dataset 3D-CNN & 31.14 & 32.33  &  19.38 &  13.30 &  10.06  \\
 \hline
   Small Dataset & 31.98 & 33.40  &  20.53 &  14.50  & 11.15  \\
 \hline
 Domain Adaptation & 34.41 & 34.84  & 22.07 & 15.75 &  12.21  \\
 \hline
 Multitask  & \bftab 36.28 & \bftab 37.22  &   \bftab 23.88 &   \bftab  17.08 &  \bftab 13.25   \\
 \hline
 Multitask + Full Frames  & \bftab 38.05 & \bftab 38.62  &   \bftab 25.26 &   \bftab  18.37 &  \bftab 14.42   \\
 \hline
\end{tabular}
\end{center}
\end{table}

\begin{figure}[htbp]
\begin{center}
\includegraphics[width=1\columnwidth]{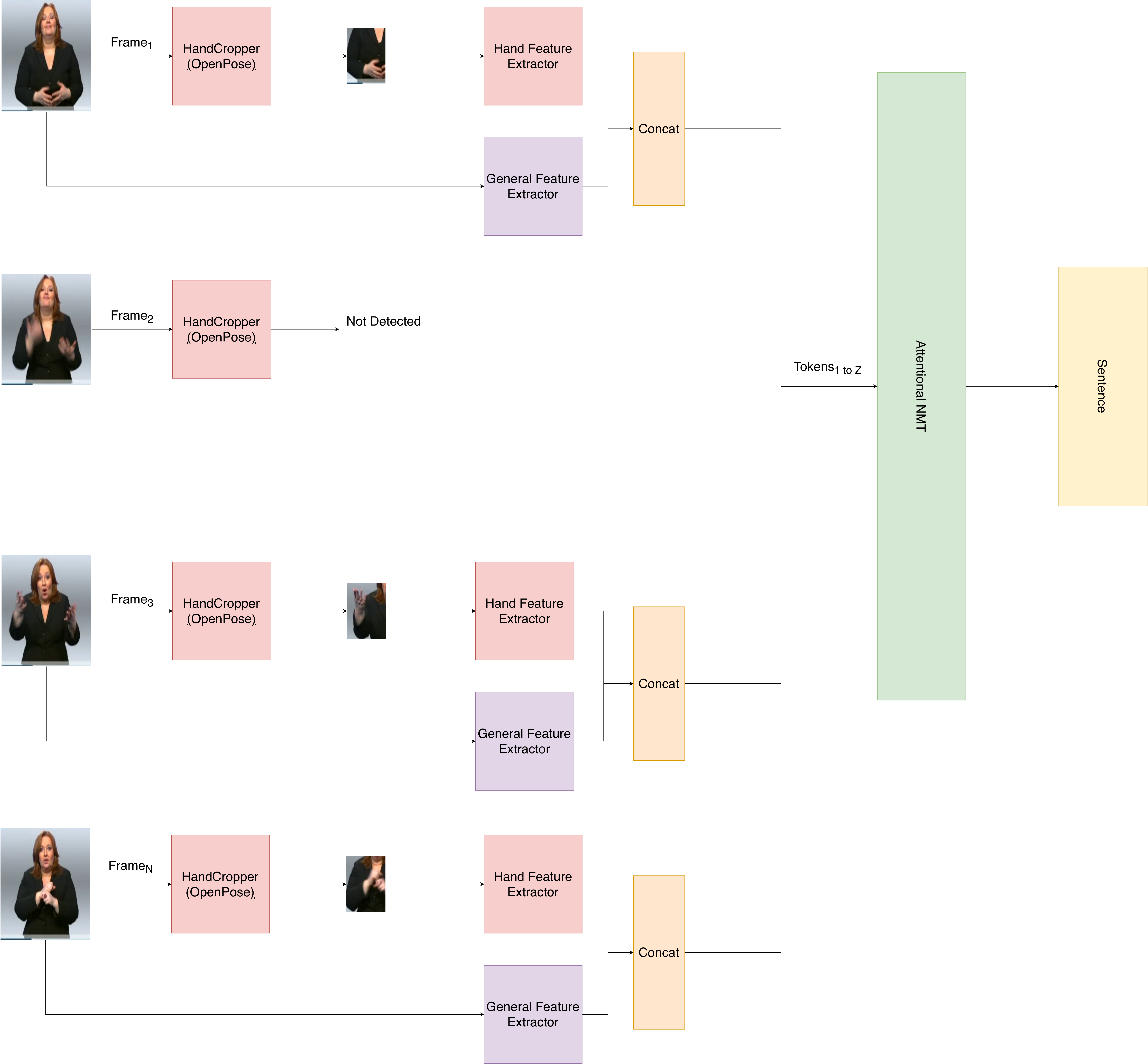}
\end{center}
\caption{Combination of right hands and full frames in Tokenization layer.}
\vskip\baselineskip
\label{fig:hand-full}
\end{figure}

\par Small dataset in Table \ref{table:hand-shape-tokenization} refers to the network trained on the small dataset. The second row indicates that a pre-trained network fine-tuned on a small dataset leads to a considerable increase in all metrics. Compared to the third rows of Table \ref{table:imagenet}, the small dataset brings additional supervision which leads to about 2.09 points increase in BLEU-4 score. The big dataset improves the scores by about 3.11 points in BLEU-4 score. Domain adaptation seems to be ineffective as it attains nearly the same results with Big Dataset though it is the most successful method in classification. It means that the sequence-to-sequence model can handle the domain shift problem implicitly. 

\par The first two rows of Table \ref{table:hand-shape-tokenization} presents the results of 2D-CNN and 3D-CNN based approaches trained on the big dataset. The 3D-CNN setting cannot attain the scores as the 2D-CNN does. There is a significant difference, 2.11 points in BLEU-4 and 2.45 points in ROUGE. A major reason may be that 3D-CNNs cannot focus on the desired hand shapes in a window of consecutive frames. In a certain context, irrelevant hand shapes before and after the desired one may be confused with other annotated hand shapes. Therefore, 3D-CNNs are not very suitable to be trained with weakly labeled hand shape images to leverage additional supervision to the tokenization layer. 

\par Multitask learning is the most successful method with 13.25 BLEU-4 score. In comparison with the third row of Table \ref{table:endtoend}, Multitask learning improves the results by 4.39 points in ROUGE and 3.00 points in BLEU-4 score. It is shown that if there is not enough data, Multitask learning is a proper way to combine weak signals to obtain a strong one. 

\par Another key conclusion is that adding small supervision over hand shape learning attains more accurate results than human keypoints. Every result in Table \ref{table:hand-shape-tokenization} is better than the third row of Table \ref{table:keypoint}. This means that OpenPose outputs are not reliable enough for hand shape representation in SL. 
\par In addition to that, we add another feature source in our tokenization layer. Up to now, it is clear that learning hand shapes is the key factor to understand signers. However, experiments in Section \ref{list:transfer-imagenet} also indicate that the general frame context may improve translation quality. Therefore, we modified our tokenization approach as seen in Figure \ref{fig:hand-full}. The hand feature extractor is Multitask and we use Resnet-34 (512) \cite{resnet} as a general feature extractor. The reason why it is chosen instead of Inception v3 (2048) is the spatio-embedding size. We want to keep the feature size as the smallest for computational concerns. The last two rows of Table \ref{table:hand-shape-tokenization} indicate that the general context improves scores by 1.77 points in ROUGE and 1.17 points in BLEU-4 scores. 

\section{Transfer Learning from Kinetics}
\label{list:transfer-kinetics}
	In this section, we conduct experiments to seek answers for two questions. Besides stationary hand shapes, action recognition data may be useful as an additional data source. Recent 3D-CNN architectures are capable of operating both in the spatial and temporal domains. In other words, 3D-CNNs can both model action and identify hand shapes. We test this assumption with I3D proposed by \cite{I3D}. This is an adapted version of Inception \cite{incep} trained on ImageNet.

\par The frame sequences are divided into eight-frame-long splits with the same order as in Figure \ref{fig:3d-cnn}. The number of resulting spatio-temporal embeddings are about one-eighth of frame number. Therefore, we expect that the sequence-to-sequence model does not suffer from long term dependencies. It also helps to reduce the required computational burden for training. For real-time applications, this tokenization approach is seen to be the most promising one.
	
\begin{table}[thbp]
\begin{center}
\caption{ Test results of Tokenization with I3D. }
\label{table:i3d}

\begin{tabular}[t]{|l||c|c|c|c|c|}
 \hline

      & ROUGE & BLEU-1 &  BLEU-2&  BLEU-3&  BLEU-4 \\
       \hline

	Full Frames  &  \bftab  29.74 &  \bftab  29.52  &   \bftab  17.09 &   \bftab  11.64 &  \bftab  8.76   \\
 \hline
 
  Both Hands & 28.64 & 28.27 &16.14  &  10.99&  8.26  \\
 \hline
 
   Right Hands & 28.00 & 28.01  &  15.78 &  10.73 & 8.09   \\
 \hline
\end{tabular}
\end{center}
\end{table}

\begin{figure}[htbp]
\begin{center}
\includegraphics[width=1\columnwidth]{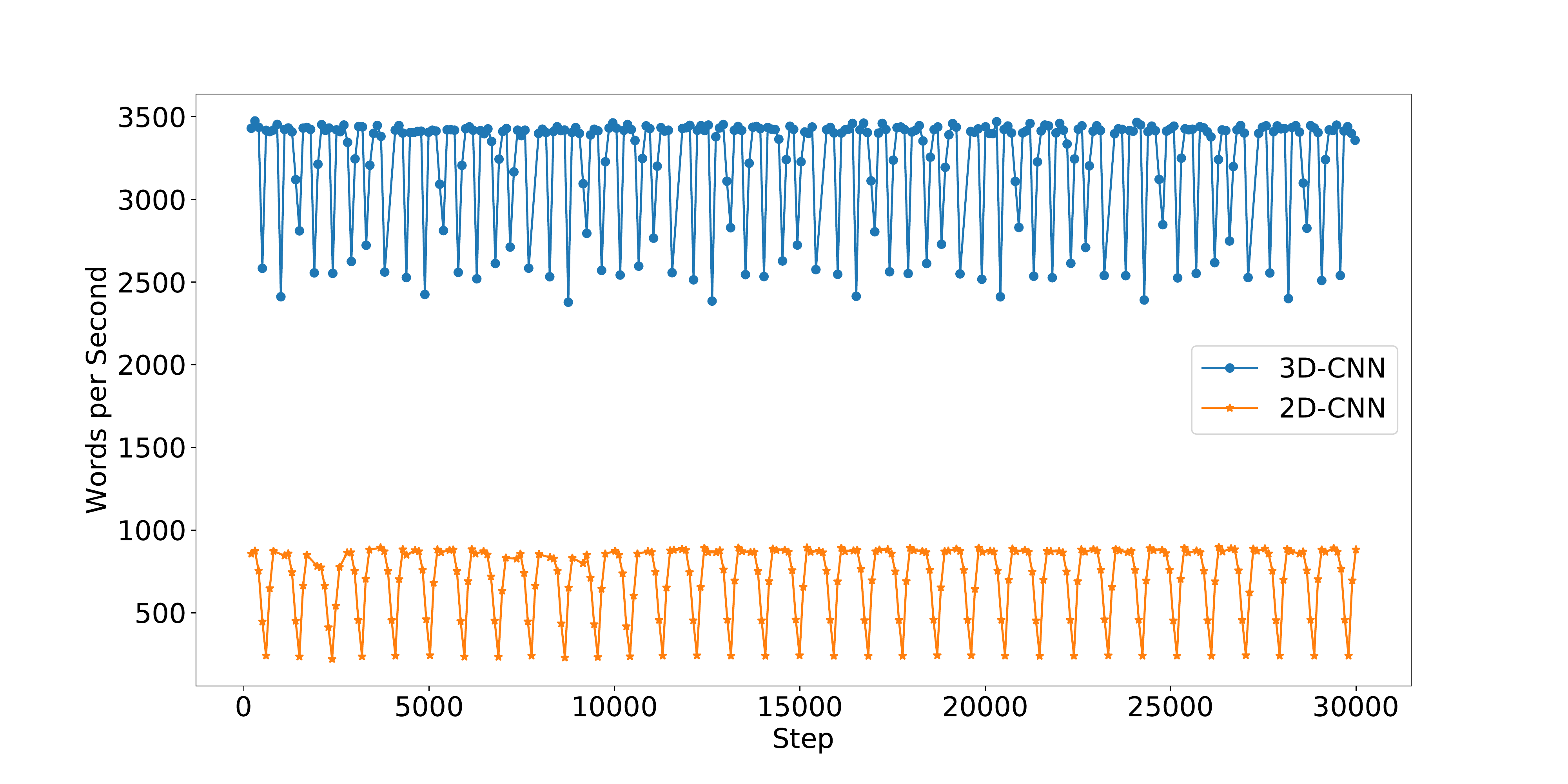}
\end{center}
\caption{Processing speed graph of NSLT with 3D-CNN and 2D-CNN.}
\vskip\baselineskip
\label{fig:3d-2d}
\end{figure}

\par This approach is comparable with the result in Table \ref{table:imagenet}. There is no certain precedence between 2D-CNNs and 3D-CNNs. They both attain the best results with full frames. 3D-CNNs are better at Rouge score (29.74) while 2D-CNNs outperform in terms of BLEU-4 score (9.4). It is clear that action recognition can be used as a related task for sign languages. Furthermore, 3D-CNNs are more efficient in terms of memory and time. As seen in Figure \ref{fig:3d-2d}, using 3D-CNN tokenization instead of 2D-CNN tokenization makes the overall model seven times faster.

\section{Tokenization Learning on Target Domain}
\label{list:target-domain}
	In this section, we include frames from  RWTH-PHOENIX-Weather 2014 to train tokenization layers. We evaluate additional supervision leveraged by our proposed approaches without target information. The big dataset now involves all of the data, German (ph), Danish and New Zealand Sign Languages. We train a baseline network on the big dataset, multitask network on the big and small dataset. The last network is DeepHand proposed by Koller \etal \cite{deephand}. This network is trained with a Hidden Markov Model that models the transition between consecutive frames to refine weak labels.  
\begin{table}[thbp]
\begin{center}
\caption{Test results with Tokenization trained on the target domain.}
\label{table:target}

\begin{tabular}[t]{|l||c|c|c|c|c|}
 \hline
   & ROUGE & BLEU-1 &  BLEU-2&  BLEU-3&  BLEU-4 \\
 \hline
  DeepHand & \bftab  38.05 & \bftab  38.50  &  \bftab 25.64 &  \bftab 18.59 &  \bftab 14.56  \\
  \hline
  Multitask & 36.35 & 37.11  &  24.10  &  17.46 &  13.50  \\
 \hline
  Baseline  & 35.22 & 35.97  & 23.10 &  16.59 &  12.89  \\
 \hline
\end{tabular}
\end{center}
\end{table}

\par As expected, DeepHand outperforms other networks as it uses all of the available information. Multitask learning is nearly the same as Multitask in Table \ref{table:hand-shape-tokenization}. Despite the fact that ph in Table \ref{table:one-million-images} is two times bigger than the combination of others, it brings little improvement in all metrics. This implies that Multitask learning can reach its maximum capacity with less amount of data. Another important finding is that Baseline scores in Table \ref{table:target} are outperformed by Multitask scores in Table \ref{table:hand-shape-tokenization}. This means that the annotation approach of the small dataset provides much more crucial abilities to the network than a huge amount of data labeled with one regime.

\section{Analysis of Results}
	In this section, we discuss whether our approach really improves the results. We compare our Multitask Learning approach with the baseline methods and DeepHand. DeepHand is a special network trained on over one million hand images involving samples from the training set of the translation dataset, RWTHPHOENIXWeather 2014T. The hand shape dataset is automatically annotated so the labels are very noisy. In \cite{deephand}, they propose a method to refine labels with sequence information. Therefore, DeepHand is not a good candidate for a generic tokenization layer as it benefits from supervision from the target domain along with sequence knowledge. Therefore, we consider this method as an upper bound for hand shape classification and frame-level tokenization with right hands. Besides that, we provide insights into how our approach increases translation quality.
	
\par The statistical significance in NMT is discussed in the work of \cite{statistical-significance}. It proposes methods to conclude a confidence score if a translation system is better than another. We use paired bootstrap resampling to validate our contributions. We use statistical validation methods as repeating experiments many times is not feasible in NSLT research. When the sample size is very limited, we use bootstrap sampling to determine a confidence interval. Paired bootstrap resampling is a method to assign confidences to our claims. NMT adopts this method to test whether a translation system is statistically significantly better than another. For a large number of times, we sample sentences with replacement to construct test sets. We evaluate the systems on these test sets and compare the results. If the first system is better than the second one in \%95 of the trials, we conclude that our claim is true with \%95 confidence. We have only 642 test sentences. Therefore, we generate three different 1000-sized subsets in which a sample consists of 250,400 and 600 sentences sampled with replacement. Eventually, we check our claims listed in Table \ref{table:ss} by this method to provide a more reliable conclusion.

\par Table \ref{table:ss} displays the confidence of the claims in the system comparison column. Baseline refers to the 2D-CNN tokenization trained on the Big Dataset without the target annotation. Also, no additional technique is applied, such as Multitask learning or Domain Adaptation. The table shows that the Multitask approach clearly outperforms the baseline, but does not attain DeepHand level results. This is to say that the representation of hand shapes can be improved without accessing target domain supervision. The target supervision is more useful if it exists. Moreover, our approach gets stronger when it is enhanced with the frame context. The last row of the table indicates that DeepHand is not better than Multitask-FullFrames setting with high confidence. \%58.80 confidence means that the results are comparable. 
\begin{table}[thbp]
\begin{center}
\caption{Paired bootstrap sampling results on different translation comparisons. Delta refers to the differences in BLEU-4 Score and 250, 400 and 600 are the number of the sentences drawn for a sample. The table displays the confidence of the claims under different sentence sizes.}
\label{table:ss}

\begin{tabular}[t]{|l||c||c|c|c|}
 \hline
 System Comparison  & Delta & 250 & 400 & 600  \\
 \hline
 Multitask better than Baseline & 1.08 & \%94.70 & \%97.0 & \%99.30 \\ 
   \hline

 Multitask+Full-Frames better than Baseline & 2.24 & \%99.60 & \%100 & \%100 \\ 
  \hline
Multitask+Full-Frames better than Multitask & 1.16 & \%93.50 & \%96.30 & \%99.10\\ 
  \hline
DeepHand better than Baseline & 2.39 & \%99.70 & \%100 & \%100 \\   \hline
DeepHand better than Multitask & 1.31 & \%93.80 & \%96.90 & \%99.10 \\   \hline

 DeepHand better than Multitask-FullFrames & 0.15 & \%55.10 & \%58.10 & \%58.80 \\ 
  \hline
\end{tabular}
\end{center}
\end{table}

	It is shown that our approach can attain the upper bound results and outperforms the baseline with high confidence. For SLT, this can be enabled by proving better sign embeddings as better word embeddings improve in NMT. We mentioned that Multitask learning increases top-1 and top-5 accuracy in hand shape classification dramatically. In NSLT, identifying more hand shapes is expected to lead to an increase in word and gloss diversity in generated translations. However, noisy embeddings may also increase the diversity where the resulting sentences become meaningless. Table \ref{table:qa} is designed to show the unique number of words whose frequency are greater than the threshold. We ignore the stop words of German while counting the words. Golden refers to the reference translation used in BLEU-4 evaluations. Multitask learning uses fewer unique words, but provides higher BLUE and ROUGE scores compared to Baseline. As expected, Baseline produces more noisy sign embeddings as it is only trained on the weakly labeled dataset. In other words, diversity is higher, but the choice of words and the placement of words are not appropriate. Multitask learning can obtain higher results with less diversity as its embeddings are very robust. When we combine its embeddings with frame contexts, the BLEU-4 scores and the unique number of words increase. To conclude, Multitask learning refines the wrong labels while training. Furthermore, the hand shape classification results show that it covers more hand shapes compared to Baseline.

\begin{table}[thbp]
\begin{center}
\caption{The table displays unique word numbers under varying frequencies in the translation systems.}
\label{table:qa}

\begin{tabular}[t]{|l||l||c|c|c|c|c|c|}
 \hline
 BLUE-4 & System  & 0 & 1 & 2 & 3 & 5 & 10  \\
 \hline
 12.17 & Baseline  & 505 & 368 & 305 & 264 & 200 & 116 \\ 
   \hline
13.25  & Multitask & 498 & 350 & 272 & 234 & 184 & 121\\ 
  \hline
14.42 &Multitask+Full-Frames & 563 & 420 & 337 & 272 & 201 & 121\\ 
  \hline
100 & Golden & 989  & 571 & 412 & 331 & 237 & 128 \\   \hline

  \hline
\end{tabular}
\end{center}
\end{table}

\section{Comparison of Frame Levels \& Gloss Level Translation}
\label{list:comparison}
	Glosses are perfect tokens for sign language translation. However, if the ultimate goal is to construct sentences instead of gloss recognition, learning glosses restricts the model. Camgoz \etal \cite{Camgz2020SignLT} suggests that tokenization learning from glosses recognition outperforms inputting perfect glosses as tokens. The findings of our study also support this claim as additional supervision in tokenization leads to dramatic increases in translation quality. 
\par Translation from golden glosses on our dataset attains 19.26 BLEU-4 score and our domain-independent tokenization approach acquires 13.25 BLEU score with only right hands. We showed that full frames are better than only right hand crops in all of the experiments. We claim that our tokenization approach can be improved much further by combining other body parts like left hands or faces. In addition to that, we show that Multitask learning is a suitable candidate to outperform the gloss level tokenization. The most important finding of this study is that we can eliminate gloss annotation with different information sources while our method is applicable to other datasets.

\chapter{CONCLUSION}
\label{chapter:conclusion}

\section{Remarks}
\label{list:remarks}
	Firstly, we show that the joint learning of tokenization and sequence-to-sequence models has certain shortcomings. As collecting a huge amount of translation samples is a very laborious task, the required additional supervision is leveraged by Sign Language Recognition tasks with the proposed tokenization learning approaches. We prepare experiments to find the most appropriate attention mechanism to deal with long sequences in frame-level tokenization. To conduct a detailed analysis, we generate two different versions of  RWTH-PHOENIX-Weather 2014T with hand crops. We propose two different tokenization learning methods and compare them with existing tokenization methods. 3D-CNNs are adapted into Sign Language Translation for the first time.
	\par The experiments are divided into two main categories, with or without access to the target domain. This division is aimed to test the effectiveness of our proposed approaches as a generic tokenization layer. Without target domain knowledge, we illustrate the positive effects of related tasks to Sign Language Translation. First, we reveal that 3D-CNNs are also good candidates since they are more efficient in terms of required computational effort and memory amount. Secondly, our proposed approach accomplishes to learn from weak signals to obtain a robust tokenization layer by Multitask learning. Moreover, domain adaptation improves classification results by mitigating domains shifts, yet does not have any effect on translation task. We infer that sequence-to-sequence models can handle domain shifts implicitly.
	\par We increase the upper-bound of frame-level tokenization. Without target knowledge, we approximate the upper-bound with our proposed approach. Our study involves four different sign languages. One of the key findings in the study emphasizes that adding a new data source may result in advancements even if they are very limited.
	\par In comparison with gloss-level tokenization, frame-level tokenization is more promising for further improvement. We close the gap between two different tokenization approaches by nearly 50\% improvement in BLEU-4 score. The development of gloss-level tokenization is restricted to tedious annotation whereas frame-level tokenization may be improved with data from other research fields.

\section{Future Work}
\label{list:future-work}
	We attain the best results with only right-hands while the experiments illustrate that other body parts may also be taken into account. Therefore, new studies may be conducted on how to fuse body parts in frame-level tokenization. We believe that this may enable further improvement in SLT.
\par Another research direction may be on modifying 3D-CNNs. They are proved to be good tokenization layers, but additional supervision may be leveraged through them. In addition to action recognition, new tasks may be established to train 3D-CNNs. Considering their ability of temporal operation, they may be even replaced with Encoder in Encoder-Decoder (ED) architecture. 

\bibliographystyle{styles/fbe_tez_v13}
\bibliography{references}

\appendix
\chapter{ATTENTIONAL NMT MODEL}
\section{Implicit Tokenization}
\begin{itemize}
	\item feature size:  4096
	\item RNN type: GRU
	\item Attention Type: Luong
	\item Layer Size: 4
	\item Hidden Unit Size: 1024
	\item Dropout Rate: 0.2
	\item Residual: True
	\item Learning Rate: 1e-5
	\item Optimizer: Adam
	\item Batch Size: 1
	\item Word Embedding Size: 256
\end{itemize}

\section{Bahdanau \& Luong Model Details}
\begin{itemize}
	\item feature size:  2048
	\item RNN type: GRU
	\item Layer Size: 4
	\item Hidden Unit Size: 512
	\item Dropout Rate: 0.2
	\item Residual: True
	\item Initial Parameter Std: 0.02
	\item Learning Rate: 1e-4
	\item Optimizer: Adam
	\item Batch Size: 16
	\item Word Embedding Size: 256
\end{itemize}
\section{Transformer Model Details}
\begin{itemize}

	\item Layer Size: 6
	\item Head Number: 8
	\item Hidden Unit Size: 1024
	\item Dropout Rate: 0.2
	\item Attentional Dropout: 0.1
	\item Relu Dropout: 0.1
	\item Residual: True
	\item Initial Parameter Std: 0.02
	\item Learning Rate: 1e-4
	\item Optimizer: Adam
	\item Batch Size: 16
	\item Word Embedding Size: 128
\end{itemize}
\chapter{CNN TRAINING}
\section{Multitask Learning}
\begin{itemize}
 \item $L_{\texttt{total}} = L_1 + \beta(t) L_2$ 
 \item $\beta(t) = 0.1$
 \item Learning Rate: 1e-4
 \item Optimizer: Adam
\end{itemize}

\section{Domain Adaptation}
\begin{itemize}
 \item $\gamma$: 2.5 (equation 14 in \cite{domainadaptation})
 \item Domain Classifier Layer Size: 2
 \item Doman Classifier Hidden Unit Size: 2048 
 \item Learning Rate: 5e-3
 \item Optimizer: SGD (Not Converged with Adam)
\end{itemize}

\end{document}